\documentclass[10pt,twocolumn,letterpaper]{article}
\pdfoutput=1
\usepackage[pagenumbers]{cvpr} 
\usepackage[pagebackref,breaklinks,colorlinks,allcolors=cvprblue]{hyperref}
\usepackage{graphicx}
\definecolor{linkcolor}{RGB}{255,0,0}
\definecolor{urlcolor}{RGB}{255,105,180}
\definecolor{citecolor}{RGB}{66,168,235}
\hypersetup{colorlinks=true,linkcolor=linkcolor,urlcolor=urlcolor}
\usepackage{pifont}
\newcommand{\cmark}{\ding{52}\xspace}%
\newcommand{\xmark}{\ding{56}\xspace}%
\definecolor{cvprblue}{rgb}{0.21,0.49,0.74}
\definecolor{Light}{rgb}{0.99, 0.92, 0.95}
\definecolor{codered}{rgb}{0.98,0.49,0.72}

\usepackage{multirow}
\usepackage{colortbl}
\title{Exploring Intrinsic Normal  Prototypes within a Single Image for\\ Universal Anomaly Detection}

\author{%
Wei Luo$^{1*}$~~~~Yunkang Cao$^{2*}$~~~~Haiming Yao$^{1*}$~~~~Xiaotian Zhang$^1$~~~~Jianan Lou$^{1}$ \\
Yuqi Cheng$^{2}$~~~~Weiming Shen$^2$~~~~Wenyong Yu$^{2\dagger}$ \\
$^1$Department of Precision Instrument, Tsinghua University\\
$^2$School of Mechanical Science \& Engineering, Huazhong University of Science \& Technology\\ %
 \small{\texttt{\{luow23, yhm22, zxt19, ljn22\}@mails.tsinghua.edu.cn}}\\
    \small{\texttt{\{cyk\_hust, yuqicheng, shenwm, ywy\}@hust.edu.cn}}
}

\begin{document}
\maketitle

\begin{abstract}
Anomaly detection (AD) is essential for industrial inspection, yet existing methods typically rely on ``comparing'' test images to normal references from a training set. However, variations in appearance and positioning often complicate the alignment of these references with the test image, limiting detection accuracy. We observe that most anomalies manifest as local variations, meaning that even within anomalous images, valuable normal information remains. We argue that this information is useful and may be more aligned with the anomalies since both the anomalies and the normal information originate from the same image. Therefore, rather than relying on external normality from the training set, we propose INP-Former, a novel method that extracts \textit{Intrinsic Normal Prototypes (INPs)} directly from the test image. Specifically, we introduce the INP Extractor, which linearly combines normal tokens to represent INPs. We further propose an INP Coherence Loss to ensure INPs can faithfully represent normality for the testing image. These INPs then guide the INP-Guided Decoder to reconstruct only normal tokens, with reconstruction errors serving as anomaly scores. Additionally, we propose a Soft Mining Loss to prioritize hard-to-optimize samples during training. INP-Former achieves state-of-the-art performance in single-class, multi-class, and few-shot AD tasks across MVTec-AD, VisA, and Real-IAD, positioning it as a versatile and universal solution for AD. Remarkably, INP-Former also demonstrates some zero-shot AD capability.  Code is available at: \href{https://github.com/luow23/INP-Former}{\textcolor{codered}{https://github.com/luow23/INP-Former}}.
\end{abstract}

\vspace{-3mm}
\section{Introduction}\label{sec:intro}

Unsupervised image anomaly detection (AD)~\cite{cao2024survey,IM-IAD} seeks to identify abnormal patterns in images and localize anomalous regions by learning solely from normal samples. This technique has seen widespread application in industrial defect detection~\cite{MVTec-AD,VisA} and medical disease screening~\cite{MVFA}. Recently, various specialized tasks have emerged in response to real-world demands, from conventional single-class AD~\cite{Patchcore, luo2024template} to more advanced few-shot AD~\cite{regad,PSNet} and multi-class AD~\cite{uniad,guo2024dinomaly, PNPT}.

\begin{figure*}[t]
    \centering
    \includegraphics[width=\linewidth]{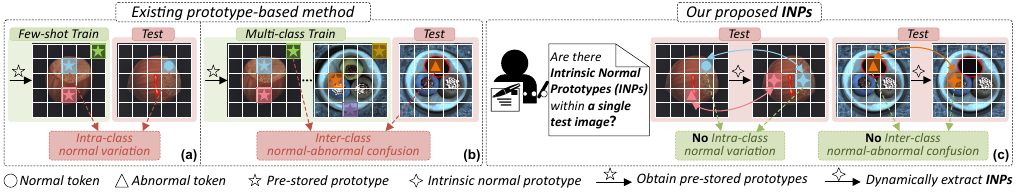}
    \caption{\textbf{Motivation for Intrinsic Normal Prototypes (\textit{INPs}).} (a) Pre-stored prototypes from few-shot normal samples may fail to represent all normal patterns. (b)  Pre-stored prototypes from one class can be similar to anomalies in another class. (c) The extracted \textbf{INPs} are concise yet well-aligned to the test image, alleviating the issues in (a) and (b).}
    \label{fig:teaser}
    \vspace{-3mm}
\end{figure*}

Although the composition of normal samples varies across these tasks, the fundamental principle remains unchanged: modeling normality in the training data and assessing whether a test image aligns with this learned normality. However, this approach can be limited due to \textit{misaligned normality} between the training data and the test image. For instance, prototype-based methods~\cite{Patchcore} extract representative normal prototypes to capture the normality of training samples. In few-shot AD, intra-class variance may lead to poorly aligned prototypes~\cite{regad}, \eg, hazelnuts in different appearances and positions, as shown in Fig.~\ref{fig:teaser}(a). Increasing the sample size can mitigate this problem but at the cost of additional prototypes and reduced inference efficiency. When there are multiple classes, \ie, multi-class AD, prototypes from one class may resemble anomalies from another, like the normal background of hazelnut is similar to the anomalies in cable in Fig.~\ref{fig:teaser}(b), leading to misclassification.


Several works have focused on extracting normality that is more aligned with the test image. For instance, some studies~\cite{regad,FOD,THFR} propose spatially aligning normality within a single class through geometrical transformations. However, spatial alignment is ineffective for certain objects, such as hazelnuts, which exhibit variations beyond spatial positions. Other approaches~\cite{PNPT,HVQ-Trans,HGAD} attempt to divide the normality in the training set into smaller, specific portions and then compare the test image to the corresponding portion of normality, but may still fail to find perfect alignment because of intra-class variances. 

Rather than attempting to extract more aligned normality from the training set, we propose addressing the issue of misaligned normality by leveraging the \textit{normality within the test image} itself as prototypes, which we term \textbf{Intrinsic Normal Prototypes (INPs)}. As illustrated in Fig.~\ref{fig:teaser}(c), normal patches within an anomalous test image can function as INPs, and anomalies can be easily detected by comparing them with these INPs. These INPs provide more concise and well-aligned prototypes to the anomalies than those learned from training data, as they typically share the same geometrical context and similar appearances with the abnormal regions within the testing image itself. Accordingly, we explore the prevalence of INPs in various AD scenarios and evaluate their potential to improve AD performance.

Although previous work~\cite{INPtexture} has attempted to utilize INP for anomaly detection, it employs handcrafted aggregated features as prototypes, thus limiting the method to zero-shot texture anomaly detection. In contrast, we introduce a learnable INP Extractor to extract normal features with adaptable shapes as INPs. We also propose an INP Coherence Loss to ensure that the extracted INPs coherently represent the normality within the test image, avoiding the capture of anomalous regions. However, some weakly representative normal regions are challenging to model with a limited set of discrete INPs, resulting in background noise (Fig.~\ref{fig:ablationlc}(c)). To address this issue, we introduce an INP-Guided Decoder, which integrates INPs into a reconstruction-based framework. This decoder leverages combinations of discrete INPs to accurately reconstruct all normal regions while effectively suppressing the reconstruction of anomalous regions, with reconstruction errors serving as anomaly scores. Furthermore, inspired by Focal Loss~\cite{focalloss} and Dinomaly~\cite{guo2024dinomaly}, we introduce a Soft Mining Loss that focuses on normal regions that are challenging to reconstruct, \ie, hard samples, thereby improving overall reconstruction quality and enhancing AD performance.

Our approach, termed INP-Former, primarily leverages vision transformers (ViTs) for both INP extraction and INP-guided reconstruction. It is worth emphasizing that INP-Former is trained exclusively on normal training images, allowing the INP extractor to learn how to extract INPs, which are dynamically derived from a single test image during the testing phase. Extensive experiments on MVTec-AD~\cite{MVTec-AD}, VisA~\cite{VisA}, and Real-IAD~\cite{real-iad} demonstrate that INP-Former achieves superior performance across multi-class, single-class, and few-shot AD tasks, \textbf{positioning INP-Former as a universal AD solution}. INP-Former also optimizes computational complexity by extracting concise INPs, \eg, images can be represented effectively using only six INPs, as shown in Sec.~\ref{sec:number_INPs}. Additionally, as demonstrated in Sec.~\ref{sec:zero-shot}, INP-Former exhibits strong generalization and can even extract INPs for unseen classes, enabling zero-shot AD capabilities.
In summary, our main contributions are:
\begin{itemize}
    \item We demonstrate that a single image can contain Intrinsic Normal Prototypes (INPs), offering concise and aligned normality for anomaly detection.
    \item We propose the INP Extractor and incorporate INPs into a reconstruction-based anomaly detection framework using the INP-Guided Decoder.
    \item We introduce the INP Coherence Loss to extract representative INPs and the Soft Mining Loss to enhance reconstruction quality.
\end{itemize}

\begin{figure*}[t]
    \centering
    \includegraphics
    [width=\textwidth]{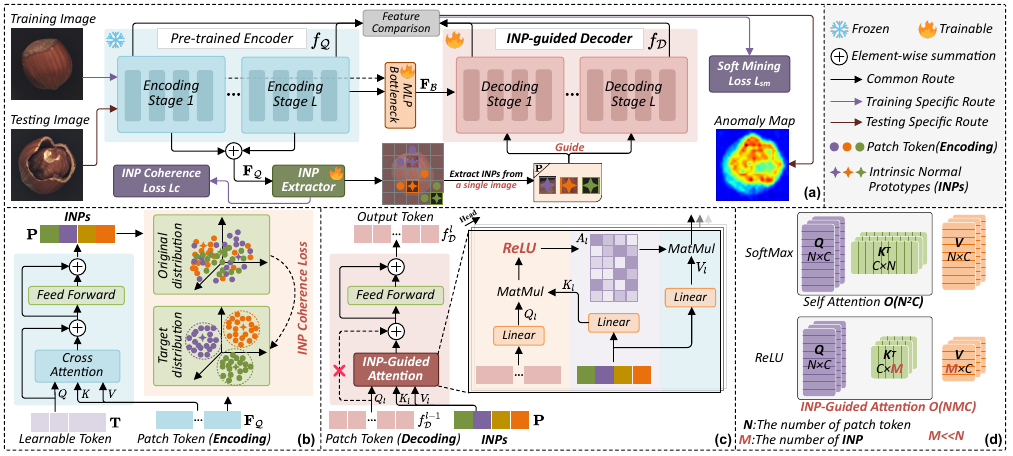}
    \caption{\textbf{Overview of our INP-Former framwork for universal anomaly detection.} (a) Our model consists of a pre-trained Encoder, an INP Extractor, a Bottleneck, and an INP-Guided decoder. The INP Extractor dynamically extracts intrinsic normal prototypes from a single image, which the INP-Guided Decoder leverages to effectively suppress anomalous features. (b) Detailed architecture of the INP Extractor. (c) Detailed architecture of each layer in the INP-Guided Decoder. (d) Comparison of computational complexity between INP-Guided Attention and Self Attention. It is important to note that the patch token \textbf{(Encoding)} and patch token \textbf{(Decoding)} refer to the patch tokens utilized during the encoding and decoding stages, respectively.}
    \label{fig:framework}
    \vspace{-3mm}
\end{figure*}
\section{Related Works}\label{sec:related_works}

\subsection{Universal Anomaly Detection}

There are numerous unsupervised AD tasks, ranging from conventional single-class AD to recent few-shot and multi-class AD setups. We refer to these collectively as universal anomaly detection.

\noindent \textbf{Single-Class Anomaly Detection:} This setup was originally introduced by MVTec-AD~\cite{MVTec-AD} and involves developing distinct AD models for each class. Typically, images are embedded into a feature space using a pre-trained encoder, after which various schemes, such as reconstruction-based~\cite{AMI-Net,RealNet, varad}, knowledge-distillation-based~\cite{RD4AD, RDplus}, prototype-based~\cite{Patchcore,FOD}, and embedding-based~\cite{liu2023simplenet, zhang2023destseg} methods, are employed to learn the normality of the given class. While these approaches achieve strong performance, their reliance on class-specific models limits scalability when dealing with a wide range of classes.

\noindent \textbf{Few-shot Anomaly Detection:} In practical scenarios, the number of available normal samples may also be limited, motivating the development of few-shot AD methods. In this case, normal samples may not fully capture the variability of normality. To address this challenge, approaches such as spatial alignment~\cite{regad} or contrastive learning~\cite{PSNet} are used to create more compact and representative normal embeddings. Recently, Vision-Language Models (VLMs) like CLIP~\cite{clip} have proven effective for few-shot AD due to their broad, pre-trained knowledge. These VLMs not only provide descriptive visual embeddings but also compute the similarity between text prompts and test images, as seen in works like WinCLIP~\cite{WinClip}, AnomalyGPT~\cite{AnomalyGPT}, and InCTRL~\cite{InCTRL}. Some approaches, such as AdaCLIP~\cite{AdaCLIP}, even enable zero-shot AD through VLMs.

\noindent \textbf{Multi-Class Anomaly Detection:} Developing separate models for each class can be resource-intensive, prompting interest in multi-class AD, also known as unified AD~\cite{uniad}, which aims to build a single model for multiple classes. UniAD~\cite{uniad} pioneered a unified reconstruction framework for anomaly detection, followed by HVQ-Trans~\cite{HVQ-Trans} addressed the identical shortcut problem using a vector quantization framework. More recent approaches, such as MambaAD~\cite{he2024mambaad} and Dinomaly~\cite{guo2024dinomaly}, further enhance multi-class AD performance by leveraging advanced models, \ie, the State Space Model Mamba~\cite{mamba} and DINO~\cite{DINOV2-R}, respectively. However, these methods lack the functionality to derive aligned normality with the test image. On the contrary, we extract INPs from the testing image, bringing aligned and precise normality for anomaly detection.

\subsection{Prototype Learning}

Prototype learning~\cite{snell2017prototypical} aims to extract representative prototypes from a given training set, which are then used for classification by measuring their distances to a test sample in a metric space. This technique is widely used in few-shot learning~\cite{li2021adaptive}. Several AD methods also employ prototype learning. For example, PatchCore~\cite{Patchcore} extracts multiple normal prototypes to represent the normality of the training data, directly computing the minimal distances to the test sample for anomaly detection. Other approaches~\cite{park2020learning,lv2021learning,huang2022pixel,MemAE} incorporate prototypes into the reconstruction process to avoid the identical shortcut issue. Specifically, they replace the original inputs with combinations of learned normal prototypes, ensuring that the inputs to the reconstruction model contain only normal elements. However, these methods rely on pre-stored normal prototypes extracted from the training set, which can suffer from the misaligned normality problem. In contrast, our INPs are dynamically extracted from the test image, providing more aligned alternatives for normality representation.

\section{Method: INP-Former}\label{sec:method}
\subsection{Overview}
To fully exploit the advantages of INPs in anomaly detection, we propose INP-Former, as depicted in Fig.~\ref{fig:framework}(a). The model dynamically extracts INPs from a single image and utilizes them to guide the feature reconstruction process, with the reconstruction errors serving as anomaly scores. Following RD4AD~\cite{RD4AD} and Dinomaly~\cite{guo2024dinomaly}, we adopt a feature reconstruction framework. Specifically, it comprises four key modules: a fixed pre-trained Encoder $\mathcal{Q}$, an INP Extractor $\mathcal{E}$, a Bottleneck $\mathcal{B}$, and an INP-Guided Decoder $\mathcal{D}$. The input image $\mathbf{I}\in\mathbb{R}^{H\times W\times 3}$ is first processed by the pre-trained Encoder $\mathcal{Q}$ to extract multi-scale latent features $f_{\mathcal{Q}}=\{f_{\mathcal{Q}}^{1}, \dots, f_{\mathcal{Q}}^{L}|f_{\mathcal{Q}}^{l}\in\mathbb{R}^{N\times C}, N=\frac{HW}{k^2}\}$, where $k$ represents the downsampling factor. Next, the INP Extractor $\mathcal{E}$ extracts $M$ INPs $\mathbf{P}=\{p_{1},\dots, p_{M}|p_{m}\in\mathbb{R}^{C}\}$ from the pre-trained features, with an INP coherence loss ensuring that the extracted INPs consistently represent normal features during testing. The Bottleneck $\mathcal{B}$ subsequently fuses the multi-scale latent features, producing the fused output $F_{\mathcal{B}}=\mathcal{B}(f_{\mathcal{Q}})$. Following the bottleneck, the extracted INPs are utilized to guide the Decoder $\mathcal{D}$ to yield reconstruction outputs $f_{\mathcal{D}}=\{f_{\mathcal{D}}^{1}, \dots, f_{\mathcal{D}}^{L}|f_{\mathcal{D}}^{l}\in\mathbb{R}^{N\times C}\}$ 
with only normal patterns, thus the reconstruction error between $f_{\mathcal{Q}}$ and $f_{\mathcal{D}}$ can serve as the anomaly score. It is worth noting that we adopt the group-to-group feature reconstruction strategy introduced in Dinomaly~\cite{guo2024dinomaly}.


\subsection{INP Extractor}
Existing prototype-based methods~\cite{Patchcore, MemAE, MNAD} store local normal features from the training data and compare them with test images. However, the misaligned normality between these pre-stored prototypes and the test images and the lack of global information lead to suboptimal detection performance. To address these limitations, we propose the INP Extractor to dynamically extract INPs with global information from the test image itself. 

Specifically, as illustrated in Fig.~\ref{fig:framework}(b), instead of extracting representative local features as done in PatchCore~\cite{Patchcore}, we employ cross attention to aggregate the global semantic information of the pre-trained features $\mathbf{F}_{\mathcal{Q}}\in\mathbb{R}^{N\times C}$ with $M$ learnable tokens $\mathbf{T}=\{t_{1},\dots, t_{M}|t_{m}\in\mathbb{R}^{C}\}$. Here $\mathbf{F}_\mathcal{Q}$ is used as the key-value pairs, while $\mathbf{T}$ serve as the query, allowing $\mathbf{T}$ to linearly aggregate $\mathbf{F}_\mathcal{Q}$ into INPs $\mathbf{P}=\{p_{1},\dots, p_{M}|p_{m}\in\mathbb{R}^{C}\}$. 
\begin{equation}
\begin{aligned}
    &\mathbf{F}_{\mathcal{Q}} = \operatorname{sum}(\{f_{\mathcal{Q}}^{1},\dots,f_{\mathcal{Q}}^{L}\})\\
    &Q = \mathbf{T}W^{Q}, K = \mathbf{F}_{\mathcal{Q}}W^{K}, V=\mathbf{F}_{\mathcal{Q}}W^{V}\\
    &\mathbf{{T}'} =\operatorname{Attention}(Q,K,V) + \mathbf{T}\\ 
    &\mathbf{P} = \operatorname{FFN}(\mathbf{{T}'}) + \mathbf{{T}'}
\end{aligned}
\end{equation}
\noindent where $\operatorname{sum}(\cdot)$ denotes the element-wise summation. $Q\in\mathbb{R}^{M\times C}$ and $K,V\in\mathbb{R}^{N\times C}$ represent the query, key and value, respectively. $W^{Q},W^{k},W^{v}\in\mathbb{R}^{C\times C}$ are the learnable projection parameters for $Q,K,V$; $\operatorname{FFN}(\cdot)$ represents the feed-forward network.

To ensure that INPs coherently represent normal features while minimizing the capture of anomalous features during the testing process, we propose an INP coherence loss $\mathcal{L}_{c}$  to minimize the distances between individual normal features and the corresponding nearest INP.
\begin{equation}
\label{eq:lc}
\begin{aligned}
    &d_{i} = \underset{m\in\{1,\dots,M\}}{\operatorname{min}}\mathcal{S}(\mathbf{F}_{\mathcal{Q}}(i),p_{m})\\
    &\mathcal{L}_{c} = \frac{1}{N}\sum_{i=1}^{N}d_{i}
\end{aligned}
\end{equation}
\noindent where $\mathcal{S}(\cdot, \cdot)$ denotes the cosine distance. $d_i$ represents the distance between the query feature $\mathbf{F}_{\mathcal{Q}}(i)$ and the corresponding nearest INP item. Fig.~\ref{fig:ablationlc} visually illustrates the effectiveness of $\mathcal{L}_{c}$.

\begin{table}[!t]
\centering
\caption{Comparison of \textbf{computational cost} and \textbf{memory usage}.}
\label{table:comparison of computational}
\fontsize{11}{14}\selectfont{
\resizebox{0.95\linewidth}{!}{
\begin{tabular}{c|cc}
\toprule[1.5pt]
           & \multicolumn{2}{c}{Number of multiplicaiton and addition}          \\ \midrule
Calculation & \multicolumn{1}{c|}{Vanilla Self Attention} & \textbf{INP-Guided Attention} \\ \midrule
$A_{l}=Q_{l}(K_{l})^{T}$      & \multicolumn{1}{c|}{943 496 960}            & \textbf{7 220 640}            \\ \midrule
${f_{\mathcal{D}}^{l-1}}'=A_{l}V_{l}$        & \multicolumn{1}{c|}{943 509 504}            & \textbf{6 623 232}            \\ \midrule
 & \multicolumn{2}{c}{Memory usage (MB)}                                         \\ \midrule
$Q_l$/$K_l$/$V_l$/$A_l$    & \multicolumn{1}{c|}{2.30/2.30/2.30/2.34}                & 2.30/\textbf{0.018}/\textbf{0.018}/\textbf{0.018}              \\ \bottomrule[1.5pt]
\end{tabular}}}
\vspace{-3mm}
\end{table}
\begin{table*}[!ht]
\centering
\caption{\textbf{Multi-class} anomaly detection performance on different AD datasets. The best in \textbf{bold}, the
second-highest is \underline{underlined}.}
\label{table:multi-class-main-performance}
\fontsize{10}{14}\selectfont{
\resizebox{\textwidth}{!}
{\begin{tabular}{c|cccccc}
\toprule[1.5pt]
Dataset~$\rightarrow$    & \multicolumn{2}{c|}{MVTec-AD~\cite{MVTec-AD}}                                                     & \multicolumn{2}{c|}{VisA~\cite{VisA}}                                                         & \multicolumn{2}{c}{Real-IAD~\cite{real-iad}}         \\ \midrule
Metric~$\rightarrow$     & \multicolumn{6}{c}{Image-level(I-AUROC/I-AP/I-F1\_max)\hspace{10mm}Pixel-level(P-AUROC/P-AP/P-F1\_max/AUPRO)}                                                                                                                        \\  \cmidrule{2-7}
Method~$\downarrow$     & Image-level    & \multicolumn{1}{c|}{Pixel-level}                                 & Image-level    & \multicolumn{1}{c|}{Pixel-level}                                 & Image-level    & Pixel-level         \\ \midrule
RD4AD~\cite{RD4AD}      & 94.6/96.5/95.2 & \multicolumn{1}{c|}{96.1/48.6/53.8/91.1}                         & 92.4/92.4/89.6 & \multicolumn{1}{c|}{98.1/38.0/42.6/91.8}                         & 82.4/79.0/73.9 & 97.3/25.0/32.7/89.6 \\
UniAD~\cite{uniad}      & 96.5/98.8/96.2 & \multicolumn{1}{c|}{96.8/43.4/49.5/90.7}                         & 88.8/90.8/85.8 & \multicolumn{1}{c|}{98.3/33.7/39.0/85.5}                         & 83.0/80.9/74.3 & 97.3/21.1/29.2/86.7 \\
SimpleNet~\cite{liu2023simplenet}   & 95.3/98.4/95.8 & \multicolumn{1}{c|}{96.9/45.9/49.7/86.5}                         & 87.2/87.0/81.8 & \multicolumn{1}{c|}{96.8/34.7/37.8/81.4}                         & 57.2/53.4/61.5 & 75.7/2.8/6.5/39.0   \\
DeSTSeg~\cite{zhang2023destseg}    & 89.2/95.5/91.6 & \multicolumn{1}{c|}{93.1/54.3/50.9/64.8}                         & 88.9/89.0/85.2 & \multicolumn{1}{c|}{96.1/39.6/43.4/67.4}                         & 82.3/79.2/73.2 & 94.6/37.9/41.7/40.6 \\
DiAD~\cite{diad}      & 97.2/99.0/96.5 & \multicolumn{1}{c|}{96.8/52.6/55.5/90.7}                         & 86.8/88.3/85.1 & \multicolumn{1}{c|}{96.0/26.1/33.0/75.2}                         & 75.6/66.4/69.9 & 88.0/2.9/7.1/58.1   \\
MambaAD~\cite{he2024mambaad}    & 98.6/99.6/97.8 & \multicolumn{1}{c|}{97.7/56.3/59.2/93.1}                         & 94.3/94.5/89.4 & \multicolumn{1}{c|}{98.5/39.4/44.0/91.0}                         & 86.3/84.6/77.0 & 98.5/33.0/38.7/90.5 \\
Dinomaly~\cite{guo2024dinomaly}   & \underline{99.6}/\underline{99.8}/\underline{99.0} & \multicolumn{1}{c|}{\underline{98.4}/\underline{69.3}/\underline{69.2}/\underline{94.8}}                         & \underline{98.7}/\underline{98.9}/\underline{96.2} & \multicolumn{1}{c|}{\underline{98.7}/\textbf{53.2}/\textbf{55.7}/\textbf{94.5}}                         & \underline{89.3}/\underline{86.8}/\underline{80.2} & \underline{98.8}/\underline{42.8}/\underline{47.1}/\underline{93.9} \\ \midrule
\rowcolor{Light}
\textbf{INP-Former} & \textbf{99.7}/\textbf{99.9}/\textbf{99.2} & \multicolumn{1}{c|}{\cellcolor{Light}\textbf{98.5}/\textbf{71.0}/\textbf{69.7}/\textbf{94.9}} & \textbf{98.9}/\textbf{99.0}/\textbf{96.6} & \multicolumn{1}{c|}{\cellcolor{Light}\textbf{98.9}/\underline{51.2}/\underline{54.7}/\underline{94.4}} & \textbf{90.5}/\textbf{88.1}/\textbf{81.5} & \textbf{99.0}/\textbf{47.5}/\textbf{50.3}/\textbf{95.0} \\ \bottomrule[1.5pt]
\end{tabular}}}
\vspace{-3mm}
\end{table*}
\subsection{INP-Guided Decoder}
While we can use the distance between testing features and their nearest INPs for anomaly detection, as illustrated in Fig.~\ref{fig:ablationlc}(c), certain low-representative normal regions are difficult to model with a limited number of discrete INPs, leading to noisy distance maps between these INPs and testing features. To address this issue, we propose the INP-Guided Decoder, aiming to reconstruct these low-representative normal regions through a combination of multiple discrete INPs and suppress the reconstruction of anomalous regions. Additionally, this decoder provides a token-wise discrepancy that can be directly leveraged for anomaly detection. As shown in Fig.~\ref{fig:framework}(c), INPs are incorporated into this decoder to guide the reconstruction process. Since INPs exclusively represent normal patterns in test images, we employ the extracted INPs as key-value pairs, ensuring that the output is a linear combination of normal INPs, thereby effectively suppressing the reconstruction of anomalous queries, \ie, the idenfical mapping issue~\cite{uniad}. Furthermore, we find that the first residual connection can directly introduce anomalous features into the subsequent reconstruction, so we remove this connection in our INP-Guided Decoder. Following the previous work~\cite{huang2024sparse}, we also employ the ReLU activation function to mitigate the influence of weak correlations and noise on the attention maps.

Mathematically, let $f_{\mathcal{D}}^{l-1}\in\mathbb{R}^{N\times C}$ denotes the output latent features from previous decoding layer. The output $f_{\mathcal{D}}^{l}\in\mathbb{R}^{N\times C}$ of the $l_{th}$ decoding layer is formulated as,
\begin{equation}
    \begin{aligned}
    &Q_{l} = f_{\mathcal{D}}^{l-1}W^{Q}_{l}, K_{l} = \mathbf{P}W^{K}_{l}, V_{l}=\mathbf{P}W^{V}_{l}\\
    &{f_{\mathcal{D}}^{l-1}}'=A_{l}V_{l}, A_{l} = \operatorname{ReLU}(Q_{l}(K_{l})^T)\\
    &f_{\mathcal{D}}^{l} = \operatorname{FFN}({f_{\mathcal{D}}^{l-1}}') + {f_{\mathcal{D}}^{l-1}}'
\end{aligned}
\end{equation}
where $Q_l\in\mathbb{R}^{N\times C}$ and $K_l, V_l\in\mathbb{R}^{M\times C}$ denote the query, key and value of the $l_{th}$ decoding layer. $W_{l}^{Q},W_{l}^{k},W_{l}^{v}\in\mathbb{R}^{C\times C}$ denote the learnable projection parameters for $Q_{l},K_{l},V_{l}$. $A_{l}\in\mathbb{R}^{N\times M}$ represent the attention map.

\noindent\textbf{Attention Complexity Analysis}: As depicted in Fig.~\ref{fig:framework}(d), the computational complexity of vanilla self-attention is $\mathcal{O}(N^2C)$, while its memory usage is $\mathcal{O}(N^2)$. In contrast, our INP-Guided Attention reduces both the computational complexity and memory usage to $\mathcal{O}(NMC)$ and $\mathcal{O}(NM)$, respectively, which can be approximated as $\mathcal{O}(NC)$ and $\mathcal{O}(N)$ due to $M\ll N$. Tab.~\ref{table:comparison of computational} offers a detailed comparison of the complexity of vanilla self-attention and INP-Guided Attention. Appendix Sec.~\textcolor{red}{C} compares the overall complexity between INP-Former and other methods. The light version of INP-Former can even be more efficient than MambaAD~\cite{he2024mambaad} yet demonstrate better performance.


\subsection{Soft Mining Loss}
Inspired by Focal Loss~\cite{focalloss} and Dinomaly~\cite{guo2024dinomaly}, different regions should be assigned varying weights based on their optimization difficulty. Accordingly, we propose Soft Mining Loss to encourage the model to focus more on difficult regions.

Intuitively, the ratio of the reconstruction error of an individual normal region to the average reconstruction error of all normal regions can serve as an indicator of optimization difficulty. Moreover, following Dinomaly~\cite{guo2024dinomaly} and ReContrast~\cite{guo2023recontrast}, we modify the feature gradients instead of applying reweighting strategies~\cite{CDO}, aiming to preserve the global structure of the feature point manifolds. Specifically, given the encoder $f_{\mathcal{Q}}^{l}$ and decoder$f_{\mathcal{D}}^{l}$ features at layer $l$, let $M^{l}$ denote the regional cosine distance. Our soft mining loss $\mathcal{L}_{sm}$ is defined as follows:
\begin{equation}
\begin{aligned}
        &w^{l}(h,w)=\left[\frac{M^{l}(h,w)}{u(M^{l})}\right]^\gamma \\
    &\mathcal{L}_{sm}=\frac{1}{L}\sum_{l=1}^{L}1-\frac{vec(f_{\mathcal{Q}}^{l})^T\cdot vec(\hat{f}_{\mathcal{D}}^{l})}{||vec(f_{\mathcal{Q}}^{l})||\,||vec(\hat{f}_{\mathcal{D}}^{l})|| }\\
     &\hat{f}_{\mathcal{D}}^{l}(h,w)=cg({f}_{\mathcal{D}}^{l}(h,w))_{w^{l}(h,w)}
\end{aligned}
\end{equation}
where $u(M^{l})$ represents the average regional cosine distance within a batch, $\gamma\geq0$ denotes the temperature hyperparameter,  $cg(\cdot)_{w^l(h,w)}$ denotes a gradient adjustment based on dynamic weight $w^{l}(h,w)$, and $vec(\cdot)$ denotes the flattening operation. The overall training loss of our INP-Former can be expressed as follows:$\mathcal{L}_{total}=\mathcal{L}_{sm}+\lambda\mathcal{L}_{c}$.


\begin{table*}[!ht]
\centering
\caption{\textbf{Few-shot (4-shot)} anomaly detection performance on different AD datasets. The best in \textbf{bold}, the
second-highest is \underline{underlined}. $\dagger$ indicates the results we reproduced using publicly available code.}
\label{table:few-shot-main-performance}
\fontsize{10}{14}\selectfont{
\resizebox{\textwidth}{!}{
\begin{tabular}{c|cc|cc|cc}
\toprule[1.5pt]
Dataset~$\rightarrow$    & \multicolumn{2}{c|}{MVTec-AD~\cite{MVTec-AD}}        & \multicolumn{2}{c|}{VisA~\cite{VisA}}              & \multicolumn{2}{c}{Real-IAD~\cite{real-iad}}                                                      \\ \midrule
Method~$\downarrow$     & Image-level    & Pixel-level         & Image-level    & Pixel-level           & \multicolumn{1}{c|}{Image-level}                            & Pixel-level         \\ \midrule
SPADE~\cite{SPADE}      & 84.8/92.5/91.5 & 92.7/-/46.2/87.0    & 81.7/83.4/82.1 & 96.6/-/43.6/87.3      & \multicolumn{1}{c|}{50.8$^{\dagger}$/45.8$^{\dagger}$/61.2$^{\dagger}$}                            & 59.5$^{\dagger}$/0.2$^{\dagger}$/0.5$^{\dagger}$/19.2$^{\dagger}$         \\
PaDiM~\cite{defard2021padim}      & 80.4/90.5/90.2 & 92.6/-/46.1/81.3    & 72.8/75.6/78.0 & 93.2/-/24.6/72.6      & \multicolumn{1}{c|}{60.3$^{\dagger}$/53.5$^{\dagger}$/64.0$^{\dagger}$}                            & 90.9$^{\dagger}$/2.1$^{\dagger}$/5.1$^{\dagger}$/67.6$^{\dagger}$         \\
PatchCore~\cite{Patchcore}  & 88.8/94.5/92.6 & 94.3/-/55.0/84.3    & 85.3/87.5/\underline{84.3} & 96.8/-/43.9/84.9      & \multicolumn{1}{c|}{66.0$^{\dagger}$/\underline{62.2}$^{\dagger}$/\underline{65.2}$^{\dagger}$}                            & 92.9$^{\dagger}$/9.8$^{\dagger}$/16.1$^{\dagger}$/68.6$^{\dagger}$         \\
WinCLIP~\cite{WinClip}    & 95.2/\underline{97.3}/\underline{94.7} & 96.2/-/\underline{59.5}/89.0    & 87.3/\underline{88.8}/84.2 & 97.2/-/\underline{47.0}/\underline{87.6}      & \multicolumn{1}{c|}{\underline{73.0}$^{\dagger}$/61.8$^{\dagger}$/61.0$^{\dagger}$}                         & \underline{93.8}$^{\dagger}$/\underline{13.3}$^{\dagger}$/\underline{21.0}$^{\dagger}$/\underline{76.4}$^{\dagger}$ \\
PromptAD~\cite{PromptAD}   & \underline{96.6}/-/-       & \underline{96.5}/-/-/\underline{90.5}       & \underline{89.1}/-/-       & \underline{97.4}/-/-/86.2         & \multicolumn{1}{c|}{59.7$^{\dagger}$/43.5$^{\dagger}$/52.9$^{\dagger}$}                         & 86.9$^{\dagger}$/8.7$^{\dagger}$/16.1$^{\dagger}$/61.9$^{\dagger}$  \\ \midrule
\rowcolor{Light} 
\textbf{INP-Former} & \textbf{97.6}/\textbf{98.6}/\textbf{97.0} & \textbf{97.0}/\textbf{65.9}/\textbf{65.6}/\textbf{92.9}& \textbf{96.4}/\textbf{96.0}/\textbf{93.0} & \textbf{97.7}/\textbf{49.3}/\textbf{54.3}/\textbf{93.1} & \multicolumn{1}{c|}{\cellcolor{Light}\textbf{76.7}/\textbf{72.3}/\textbf{71.7}} & \textbf{97.3}/\textbf{32.2}/\textbf{36.7}/\textbf{89.0} \\ \bottomrule[1.5pt]
\end{tabular}}}
\vspace{-3mm}
\end{table*}
\begin{figure*}[!ht]
    \centering
    \includegraphics[width=\linewidth]{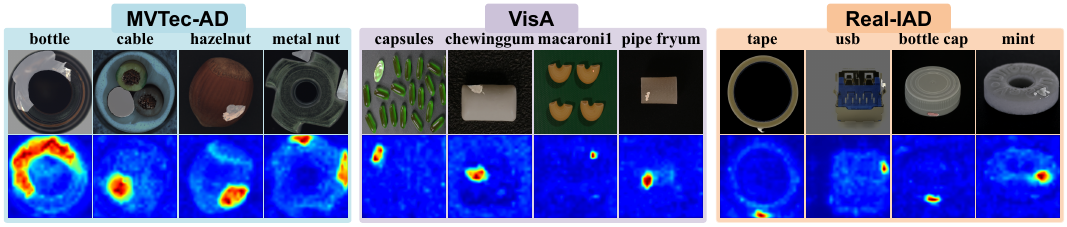}
    \caption{\textbf{Qualitative results of anomaly localization} on the MVTec-AD~\cite{MVTec-AD}, VisA~\cite{VisA}, and Real-IAD~\cite{real-iad} datasets for \textbf{multi-class anomaly detection}. The first row presents the input images with their ground truth, while the second row displays the corresponding anomaly maps.}
    \label{fig:anomalymap}
    \vspace{-3mm}
\end{figure*}

\section{Experiments}\label{sec:experiments}

\subsection{Experimental Settings}

\noindent\textbf{Datasets:}
We conduct a comprehensive analysis of the proposed INP-Former on three widely used AD datasets: \textbf{MVTec-AD}~\cite{MVTec-AD}, \textbf{VisA}~\cite{VisA}, and \textbf{Real-IAD}~\cite{real-iad}. \textbf{MVTec-AD} consists of 15 categories, with 3,629 normal images for training, and 1,982 anomalous images along with 498 normal images for testing. \textbf{VisA} contains 12 object categories, with 8,659 normal images for training and 962 normal images along with 1,200 anomalous images for testing. \textbf{Real-IAD} includes 30 different objects, with 36,645 normal images for training and 63,256 normal images along with 51,329 anomalous images for testing.

\noindent\textbf{Metrics:}
Following existing works~\cite{he2024mambaad, guo2024dinomaly}, we use the Area Under the Receiver Operating Characteristic Curve (AUROC), Average Precision (AP), and F1-score-max (F1\_max) to evaluate anomaly detection and localization. For anomaly localization specifically, we use Area Under the Per-Region-Overlap (AUPRO) as an additional metric.

\noindent\textbf{Implementation Details:}
INP-Former adopts ViT-Base/14 with DINO2-R~\cite{DINOV2-R} weights as the default pre-trained encoder. The number $M$ of INPs is set to six by default. The INP Extractor includes a standard Vision Transformer block. The layer number of the INP-Guided decoder is eight. All input images are resized to $448^2$ and then center-cropped to $392^2$. The hyperparameters $\gamma$ and $\lambda$ are set to 3.0 and 0.2, respectively. We use the StableAdamW~\cite{wortsman2023stable} optimizer with a learning rate $1e^{-3}$ and a weight decay of $1e^{-4}$ for 200 epochs. Notably, the above hyperparameters do not require any adjustment across the three datasets. Appendix Sec.~\textcolor{red}{A} presents more implementation details. Appendix Sec.~\textcolor{red}{D},~\textcolor{red}{E}, and~\textcolor{red}{F} analyze the influence of input resolution, ViT architecture, and $\lambda$, respectively.

\begin{table*}[!ht]
\centering
\caption{\textbf{Single class} anomaly detection performance on different AD datasets. The best in \textbf{bold}, the
second-highest is \underline{underlined}.}
\label{table:single-class-main-performance}
\fontsize{11.5}{14}\selectfont{
\resizebox{.825\textwidth}{!}{
\begin{tabular}{c|ccc|ccc|ccc}
\toprule[1.5pt]
Dataset~$\rightarrow$     & \multicolumn{3}{c|}{MVTec-AD~\cite{MVTec-AD}} & \multicolumn{3}{c|}{VisA~\cite{VisA}} & \multicolumn{3}{c}{Real-IAD~\cite{real-iad}} \\ \midrule
Method~$\downarrow$      & I-AUROC    & P-AP   & AUPRO   & I-AUROC  & P-AP  & AUPRO  & I-AUROC   & P-AP   & AUPRO   \\ \midrule
PatchCore~\cite{Patchcore}   & 99.1       & 56.1   & 93.5    & 95.1     & 40.1  & 91.2   & 89.4      & -      & 91.5    \\
RD4AD~\cite{RD4AD}       & 98.5       & 58.0     & 93.9    & 96.0       & 27.7  & 70.9   & 87.1      & -      & 93.8    \\
SimpleNet~\cite{liu2023simplenet}   & \underline{99.6}       & 54.8   & 90.0      & 96.8     & 36.3  & 88.7   & 88.5      & -      & 84.6    \\
Dinomaly~\cite{guo2024dinomaly}    & \textbf{99.7}       & \underline{68.9}   & \underline{95.0}      & \textbf{98.9}     & \textbf{50.7}  & \textbf{95.1}   & \underline{92.0}        & \underline{45.2}   & \underline{95.1}    \\ \midrule
\rowcolor{Light} 
\textbf{INP-Former}  & \textbf{99.7}       & \textbf{70.2}   & \textbf{95.4}    & \underline{98.5}        & \underline{49.2}     & \underline{93.8}      & \textbf{92.1}      & \textbf{48.1}   & \textbf{95.6}    \\ \bottomrule[1.5pt]
\end{tabular}}
}
\vspace{-3mm}
\end{table*}
\begin{table*}[!t]
\centering
\caption{\textbf{Overall ablation} on MVTec-AD~\cite{MVTec-AD} and VisA~\cite{VisA} datasets. \textbf{\textit{``INP''}} refers to the use of INP Extractor and INP-Guided Decoder.}
\label{table:ablation}
\fontsize{10}{14}\selectfont{
\resizebox{0.91\linewidth}{!}{
\begin{tabular}{c|ccc|cc|cc}
\toprule[1.5pt]
                    & \multicolumn{3}{c|}{Dataset~$\rightarrow$ }                                                      & \multicolumn{2}{c|}{MVTec-AD~\cite{MVTec-AD}}                                                               & \multicolumn{2}{c}{VisA~\cite{VisA}}                                                                    \\ \midrule
                   & \textbf{\textit{``INP''}}                        & $\mathcal{L}_c$                        & $\mathcal{L}_{sm}$                        & Image-level                               & Pixel-level                                     & Image-level                               & Pixel-level                                     \\ \midrule
                    & \textcolor{gray}{\xmark}                         & \textcolor{gray}{\xmark}                         & \textcolor{gray}{\xmark}                         & 98.59/99.18/97.63                         & 97.19/61.73/62.94/92.73                         & 96.58/97.18/92.89                         & 97.50/47.24/51.90/82.85                         \\
                    & \cmark                         & \textcolor{gray}{\xmark}                         & \textcolor{gray}{\xmark}                         & 99.53/99.80/98.81                         & 98.32/69.82/69.38/94.69                         & 98.11/98.23/95.22                         & 98.41/50.34/54.23/93.63                         \\
                    & \cmark                         & \cmark                         & \textcolor{gray}{\xmark}                        & 99.61/99.83/99.02                         & 98.39/70.01/69.53/\textbf{95.10}                         & 98.16/98.30/95.47                         & 98.46/51.09/54.46/93.71                         \\
\multirow{-4}{*}{\rotatebox{90}{\textbf{Module}}} & \cellcolor{Light}\cmark & \cellcolor{Light}\cmark & \cellcolor{Light}\cmark & \cellcolor{Light}\textbf{99.67}/\textbf{99.88}/\textbf{99.20} & \cellcolor{Light}\textbf{98.48}/\textbf{71.02}/\textbf{69.65}/94.87 & \cellcolor{Light}\textbf{98.90}/\textbf{99.02}/\textbf{96.57} & \cellcolor{Light}\textbf{98.90}/\textbf{51.22}/\textbf{54.74}/\textbf{94.36} \\ \bottomrule[1.5pt]
\end{tabular}}}
\vspace{-3mm}
\end{table*}

\subsection{Main Results}
\subsubsection{Multi-Class Anomaly Detection}
We compare the proposed INP-Former with several state-of-the-art (SOTA) methods for multi-class anomaly detection, including reconstruction-based methods RD4AD~\cite{RD4AD}, UniAD~\cite{uniad}, DiAD~\cite{diad}, MambaAD~\cite{he2024mambaad}, and Dinomaly~\cite{guo2024dinomaly}, and embedding-based methods SimpleNet~\cite{liu2023simplenet} and DeSTSeg~\cite{zhang2023destseg}. A detailed introduction to the comparison methods can be found in Appendix Sec.~\textcolor{red}{B}.

The experimental results on the three AD datasets are presented in Tab.~\ref{table:multi-class-main-performance}. On the widely used MVTec-AD dataset, our method achieves SOTA performance, with image-level metrics of \textbf{99.7}/\textbf{99.9}/\textbf{99.2} and pixel-level metrics of \textbf{98.5}/\textbf{71.0}/\textbf{69.7}/\textbf{94.9}. On VisA, our method achieves competitive results, attaining the best image-level metrics of \textbf{98.9}/\textbf{99.0}/\textbf{96.6}, and achieving the best or second-best pixel-level performance of \textbf{98.8}/\underline{51.2}/\underline{54.7}/\underline{94.4}. On the more complex and challenging Real-IAD dataset, our method reaches new SOTA performance, with image-level metrics of \textbf{90.5}/\textbf{88.1}/\textbf{81.5} and pixel-level metrics of \textbf{99.0}/\textbf{47.5}/\textbf{50.3}/\textbf{95.0}. Compared to the second-best results, our method improves by \textcolor{purple}{1.2$\uparrow$}/\textcolor{purple}{1.3$\uparrow$}/\textcolor{purple}{1.3$\uparrow$} at the image level and by \textcolor{purple}{0.2$\uparrow$}/\textcolor{purple}{4.7$\uparrow$}/\textcolor{purple}{3.2$\uparrow$}/\textcolor{purple}{1.1$\uparrow$} at the pixel level. The SOTA performance achieved across the three datasets showcases the effectiveness and robustness of our method. The per-class performance metrics are presented in Appendix Sec.~\textcolor{red}{H}. Appendix Sec.~\textcolor{red}{G} presents the performance of INP-Former in a more challenging scenario, which we call super-multi-class anomaly detection, thus training INP-Former on several datasets -- MVTec-AD, VisA, and Real-IAD -- simultaneously. Results show that our method can even detect anomalies in more classes without significant performance degradation. Fig.~\ref{fig:anomalymap} demonstrates the precise anomaly localization capability of our method. More qualitative results are presented in Appendix Sec.~\textcolor{red}{L}.

\begin{figure}[!t]
    \centering
\includegraphics[width=\linewidth]{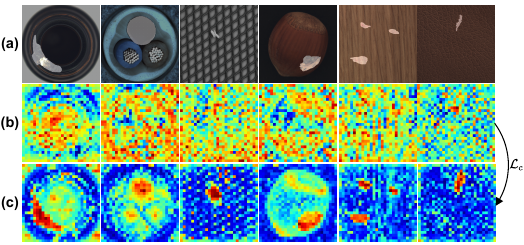}
    \caption{\textbf{Visualization of the impact of INP coherence loss $\mathcal{L}_{c}$}. (a) Input anomalous image and ground truth. (b) Distance map \textbf{without $\mathcal{L}_{c}$}. (c) Distance map \textbf{with $\mathcal{L}_{c}$}. The distance map is obtained by calculating the distance between the input features and their nearest INP terms, as described in Eq.~\ref{eq:lc}.}
    \label{fig:ablationlc}
\end{figure}

\subsubsection{Few-Shot Anomaly Detection}
We compare our method with several SOTA approaches for few-shot anomaly detection, including prototype-based methods SPADE~\cite{SPADE}, PaDiM~\cite{defard2021padim}, and PatchCore~\cite{Patchcore} and recent advances that utilize VLMs, \ie, WinCLIP~\cite{WinClip} and PromptAD~\cite{PromptAD}.

As shown in Tab.~\ref{table:few-shot-main-performance}, our method significantly outperforms previous SOTAs on three different AD datasets. Compared to the second-best results, our method achieves improvements of \textcolor{purple}{1.0$\uparrow$}/\textcolor{purple}{1.3$\uparrow$}/\textcolor{purple}{2.3$\uparrow$} in image-level scores and \textcolor{purple}{0.5$\uparrow$}/-/\textcolor{purple}{6.1$\uparrow$}/\textcolor{purple}{2.4$\uparrow$} in pixel-level scores on the MVTec-AD dataset. It similarly outperforms the second-best results on the VisA dataset, with enhancements of \textcolor{purple}{7.3$\uparrow$}/\textcolor{purple}{7.2$\uparrow$}/\textcolor{purple}{8.7$\uparrow$} for image-level and \textcolor{purple}{0.3$\uparrow$}/-/\textcolor{purple}{7.3$\uparrow$}/\textcolor{purple}{5.5$\uparrow$} for pixel-level scores. Additionally, on the Real-IAD dataset, our method surpasses the second-best results by \textcolor{purple}{3.7$\uparrow$}/\textcolor{purple}{10.1$\uparrow$}/\textcolor{purple}{6.5$\uparrow$} in image-level and \textcolor{purple}{3.5$\uparrow$}/\textcolor{purple}{18.9$\uparrow$}/\textcolor{purple}{15.7$\uparrow$}/\textcolor{purple}{12.6$\uparrow$} in pixel-level scores. The superior performance of our method in few-shot anomaly detection stems from its ability to extract INPs from a single image, eliminating the need for extensive normal data to pre-store prototypes. More comparison results on 1-shot and 2-shot are presented in Appendix Sec.~\textcolor{red}{I}.

\vspace{-3mm}
\subsubsection{Single-Class Anomaly Detection}

We further compared our proposed INP-Former with current SOTA methods for single-class anomaly detection, as shown in Tab.~\ref{table:single-class-main-performance}. The results indicate that INP-Former achieves new SOTA performance on the MVTec-AD and Real-IAD datasets and demonstrates competitive performance on the VisA dataset. Per-category performance of INP-Former is presented in Appendix Sec.~\textcolor{red}{J}.

\subsection{Ablation Study}
\subsubsection{Overall Ablation}
\label{sec-overall-ablation}

\begin{figure}[!t]
    \centering
\includegraphics[width=\linewidth]{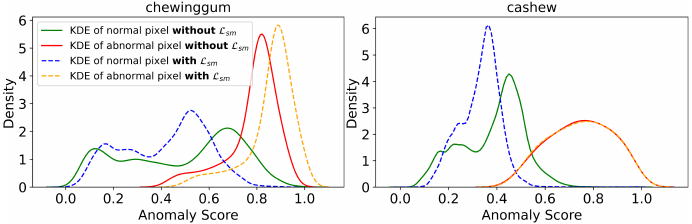}
    \caption{\textbf{Visualization of the impact of soft mining loss $\mathcal{L}_{sm}$}. We plot the Kernel Density Estimation (KDE) for the chewinggum and cashew categories in the VisA~\cite{VisA} dataset to estimate the probability density of the anomaly scores.}
    \label{fig:ablationLsm}
    \vspace{-3mm}
\end{figure}

\begin{figure}[!t]
    \centering
\includegraphics[width=\linewidth]{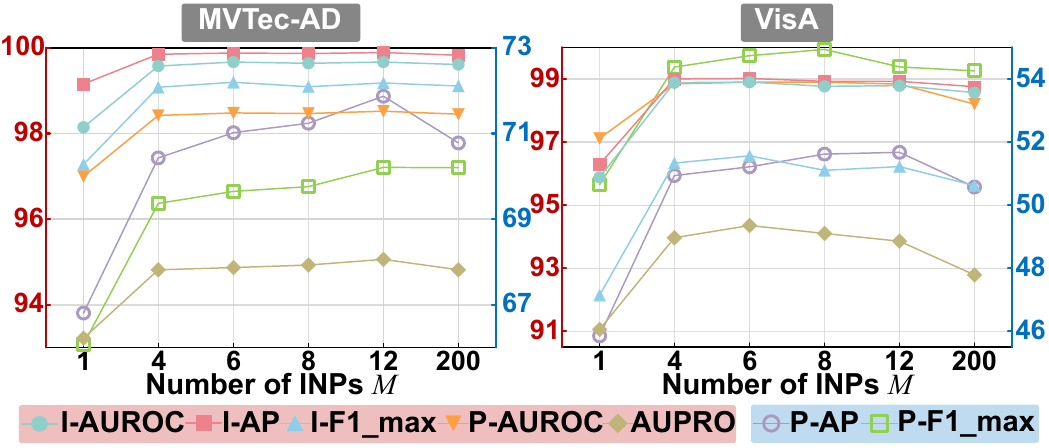}
    \caption{\textbf{Influence of the number of INPs} $M$ on model performance across the MVTec-AD~\cite{MVTec-AD} and VisA~\cite{VisA} datasets. Pixel-level AP and F1\_max use the \textcolor{blue}{right vertical axis}, while the other metrics share the \textcolor{red}{left vertical axis}.}
    \label{fig:ablation-M}
    \vspace{-5mm}
\end{figure}

\vspace{-2mm}
As shown in Tab.~\ref{table:ablation}, we conduct comprehensive experiments on MVTec-AD~\cite{MVTec-AD} and VisA~\cite{VisA} to validate the effectiveness of the proposed components, \ie, INP Extractor and INP-Guided Decoder (\textbf{\textit{``INP''}}), INP Coherence Loss ($\mathcal{L}_c$), and Soft Mining Loss ($\mathcal{L}_{sm}$). In the first row, we train a baseline model without incorporating any proposed module, {similar to the RD4AD~\cite{RD4AD} framework}. The results in the second row demonstrate that \textbf{\textit{``INP''}} significantly enhances overall performance. This improvement arises from the fact that \textbf{\textit{``INP''}} introduces an information bottleneck, {which effectively helps the model preserve normal features while filtering out anomalous ones.} The results in the third row indicate that $\mathcal{L}_{c}$ enhances the model's performance. This improvement stems from $\mathcal{L}_{c}$ ensuring that the extracted INP coherently represents normal patterns, thereby avoiding the capture of anomalous ones and establishing a solid foundation for the subsequent suppression of anomalous feature reconstruction. Fig.~\ref{fig:ablationlc} provides a more intuitive demonstration of the effectiveness of $\mathcal{L}_c$. The last row indicates that $\mathcal{L}_{sm}$ boosts overall performance, as $\mathcal{L}_{sm}$ directs the model's attention toward more challenging regions, thereby unlocking its optimal performance. Fig.~\ref{fig:ablationLsm} visually illustrates the impact of $\mathcal{L}_{sm}$, from which we can see that $\mathcal{L}_{sm}$ contributes to a smaller overlap between anomaly score distributions of normal and abnormal pixels.

\subsubsection{Influence of the Number of INPs}\label{sec:number_INPs}
As shown in Fig.~\ref{fig:ablation-M}, we conduct an ablation analysis on the number $M$ of INPs. The experimental results indicate that when $M$ exceeds four, the model's performance stabilizes. However, if $M$ becomes excessively large, the extracted INPs may also comprise information from abnormal tokens, leading to a slight decline in overall performance. In our study, we set 
$M$ to six.

\subsection{Exploration on INPs}

\subsubsection{Visualizations of INPs}
\begin{figure}
    \centering
    \includegraphics[width=1\linewidth]{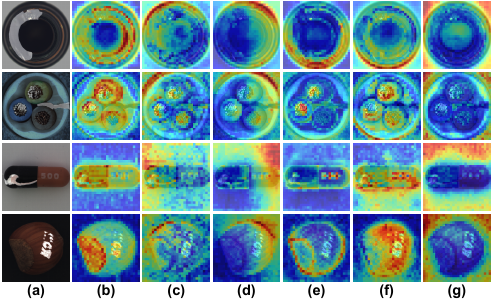}
    \caption{\textbf{Visualizations of INPs}. (a) Input anomalous image and ground truth. (b)-(g) Attention maps of six different INPs.}
    \label{fig:prototype_vis}
    \vspace{-3mm}
\end{figure}
As shown in Fig.~\ref{fig:prototype_vis}, INPs effectively capture different semantic information. Specifically, the learned INPs focus on various regions of the image, including object areas (Fig.~\ref{fig:prototype_vis}(b), (e) and (f)), object edges (Fig.~\ref{fig:prototype_vis}(c) and (d)), and background areas (Fig.~\ref{fig:prototype_vis}(g)). This diversity is attributed to our design of guiding the reconstruction process with INPs. Additionally, INP coherence loss ensures consistency in representing normal features, allowing INPs to concentrate solely on normal regions while ignoring anomalies. This mechanism ensures the decoder reconstructs features containing only normal patterns, thereby improving anomaly detection performance.



%

\subsubsection{Generalization capabilities of INP Extractor}\label{sec:zero-shot}
\begin{figure}
    \centering
    \includegraphics[width=1\linewidth]{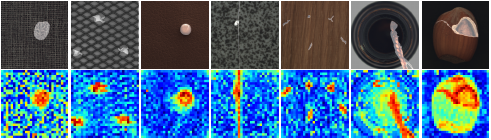}
    \caption{\textbf{Zero-shot anomaly detection results}. Here INP-Former is trained on Real-IAD~\cite{real-iad} and tested on MVTec-AD~\cite{MVTec-AD}. Distance maps are visualized.  }
    \label{fig:zero-shot}
    \vspace{-5mm}
\end{figure}
As shown in Fig. \ref{fig:zero-shot}, the INP Extractor trained on the Real-IAD \cite{real-iad} dataset is capable of detecting INPs on the unseen MVTec-AD~\cite{MVTec-AD} dataset, and the distance maps to these INPs can serve for zero-shot anomaly detection. This effectively demonstrates the INP Extractor's ability to dynamically extract INPs from a single image, with the INP coherence loss $\mathcal{L}_c$ ensuring that the extracted INPs coherently capture normal patterns. As shown in Appendix Sec.~\textcolor{red}{K}, without any specific training for zero-shot anomaly detection, our method can even outperform a specified method WinCLIP~\cite{WinClip}, achieving 88.0 and 88.7 pixel-level AUROCs on MVTec-AD and VisA, respectively.

\section{Conclusion}\label{sec:conclusion}
We propose INP-Former, a novel method for anomaly detection that explores the role of INPs. By learning to linearly combine normal tokens into INPs and using these INPs to guide the reconstruction of normal tokens, INP-Former significantly enhances anomaly detection performance. The introduction of the INP Coherence Loss and Soft Mining Loss further refines INP quality and optimizes the training process. Extensive experiments on MVTec-AD, VisA, and Real-IAD datasets demonstrate that INP-Former achieves SOTA or comparable performance across single-class, multi-class, and few-shot anomaly detection tasks. These results validate the existence and effectiveness of INPs, which can even be extracted from images in unseen categories, enabling zero-shot anomaly detection.

\noindent \textbf{Limitations \& Future Works.} 
Our method encounters certain limitations when detecting logical anomalies that closely resemble the background distribution, such as the misplaced anomalies in the Transistor class of the MVTec-AD dataset. This issue primarily arises because the misplaced anomaly in Transistor is highly similar to the background, causing INP Extractor to incorrectly extract this anomaly as INPs. A more detailed discussion can be found in Appendix Sec.~\textcolor{red}{M}. In future work, we plan to combine the proposed INPs with pre-stored prototypes to address this limitation. While the pre-stored prototypes encapsulate comprehensive semantic information, the INPs exhibit strong alignment. This integration is expected to significantly improve the model’s ability to detect logical anomalies that are similar to the background.

\noindent \textbf{Broader Impact.} This study marks the first proposal of a universal anomaly detection method that achieves exceptional performance across single-class, multi-class, and few-shot anomaly detection settings, laying a foundation for future research in general-purpose anomaly detection.

{
    \small
    \bibliographystyle{ieeenat_fullname}
    \bibliography{main}
}
\clearpage
\renewcommand{\thefigure}{S\arabic{figure}} 
\setcounter{figure}{0} 
\renewcommand{\thetable}{S\arabic{table}} 
\setcounter{table}{0} 
\appendix

\section*{Appendix}\label{sec:appendix}

\section*{Overview}
The supplementary material presents the following sections to strengthen the main manuscript:

\begin{itemize}[label=---, left=1.5em]
    \item \textbf{Sec.~\ref{sec:appen_more_implementation_details}} shows more implementation details.
    \item \textbf{Sec.~\ref{sec:appen_more_comparison_details}} shows more details about comparison methods.
    \item \textbf{Sec.~\ref{sec:appen_complexity_comparisons}} shows the complexity comparisons.
    \item \textbf{Sec.~\ref{sec:appen_influence_of_resolution}} shows the influence of input resolution.
    \item \textbf{Sec.~\ref{sec:appen_influence_of_backbone}} shows the influence of ViT architecture.
    \item \textbf{Sec.~\ref{sec:appen_influence_of_loss_weight}} shows the influence of the weight of loss functions.
    \item \textbf{Sec.~\ref{sec:appen_super_multi_class_anomaly_detection_results}} shows the super-multi-class anomaly detection.
    \item \textbf{Sec.~\ref{sec:appen_per_class_multi_class_anomaly_detection_results}} shows per-class multi-class anomaly detection results.
    \item \textbf{Sec.~\ref{sec:appen_per_class_few_shot_anomaly_detection_results}} shows more few-shot anomaly detection results.
    \item \textbf{Sec.~\ref{sec:appen_per_class_single_class_anomaly_detection_results}} shows per-class single-class anomaly detection results.
    \item \textbf{Sec.~\ref{sec:appen_per_class_zero_shot_anomaly_detection_results}} shows more zero-shot anomaly detection results.
    \item \textbf{Sec.~\ref{sec:appen_more_qualitative_results}} shows more visualized anomaly localization results
    \item 
    \textbf{Sec.~\ref{sec:appen_more_limitations}} shows a more detailed analysis of the limitations
        \item 
    \textbf{Sec.~\ref{sec:compared_with_r1}} shows a comparison of INP with handcrafted aggregated prototypes
        \item 
    \textbf{Sec.~\ref{sec:compared_with_Musc}} shows a comparison of INP with MuSc
            \item 
    \textbf{Sec.~\ref{sec:assumption}} shows more INP visualization results
\end{itemize}

\section{More implementation details}\label{sec:appen_more_implementation_details}
Building on {Dinomaly~\cite{guo2024dinomaly}}, we adopt a group-to-group supervision approach by summing the features of the layers of interest to form distinct groups. In our study, we define two groups: the features from layers 3 to 6 of ViT-Base~\cite{ViT} constitute one group, while those from layers 7 to 10 form another. We construct the anomaly detection map using the regional cosine distance~\cite{RD4AD} between the feature groups of the encoder and decoder, computing the average of the top 1\% of this map as the image-level anomaly score. In the few-shot setting, we employ data augmentation techniques similar to RegAD~\cite{regad}. Additionally, it is worth noting that in our few-shot experiments on the Real-IAD~\cite{real-iad} dataset, the term ``shot" refers to the number of images rather than the number of views. The experimental code is implemented in Python 3.8 and PyTorch 2.0.0 (CUDA 11.8) and runs on an NVIDIA GeForce RTX 4090 GPU (24GB).

\begin{table}[!t]
\centering
\caption{\textbf{Comparison of computational efficiency among SOTA methods}. mAD represents the average value of seven metrics on the Real-IAD~\cite{real-iad} dataset. The \textbf{INP-Former-S} denotes a model variant based on the ViT-Small architecture, while \textbf{INP-Former-S$^*$} refers to a model variant using the ViT-Small architexture with an image size of R256$^2$-C224$^2$.}
\label{table:complexity}
\fontsize{10}{14}\selectfont{
\resizebox{0.9\linewidth}{!}{
\begin{tabular}{c|ccc}
\toprule[1.5pt]
Method              & Params(M)     & FLOPs(G)     & mAD           \\ \midrule
RD4AD~\cite{RD4AD}               & 150.6          & 38.9         & 68.6          \\
UniAD~\cite{uniad}               & \textbf{24.5} & \textbf{3.6} & 67.5          \\
SimpleNet~\cite{liu2023simplenet}           & 72.8          & 16.1         & 42.3          \\
DeSTSeg~\cite{zhang2023destseg}             & 35.2          & 122.7        & 64.2          \\
DiAD~\cite{diad}                & 1331.3        & 451.5        & 52.6          \\
MambaAD~\cite{he2024mambaad}             & \underline{25.7}          & 8.3          & 72.7          \\
Dinomaly~\cite{guo2024dinomaly}            & 132.8         & 104.7        & 77.0            \\ \midrule
\rowcolor{Light} 
\textbf{INP-Former} & 139.8         & 98.0           & \textbf{78.8} \\ 
\textbf{INP-Former-S} & 35.1         & 24.6           & \underline{78.4} \\ 
\textbf{INP-Former-S$^{*}$} & 35.1        & \underline{8.1}           & 73.8 \\ 
\bottomrule[1.5pt]
\end{tabular}}}
\end{table}

\begin{table*}[!ht]
\centering
\caption{Influence of the \textbf{Image Size} on model performance for the MVTec-AD~\cite{MVTec-AD} dataset. R256$^2$-C224$^2$ denotes resizing the image to 256$\times$256, followed by a center crop to 224$\times$224.}
\label{table:image_size}
\fontsize{11.5}{14}\selectfont{
\resizebox{0.85\linewidth}{!}{
\begin{tabular}{c|ccc|cccc|cc}
\toprule[1.5pt]
Metric~$\rightarrow$     & \multicolumn{3}{c|}{Image-level} & \multicolumn{4}{c|}{Pixel-level} & \multicolumn{2}{c}{Efficiency} \\ \cmidrule{2-10} 
Image Size~$\downarrow$ & AUROC     & AP      & F1\_max    & AUROC  & AP    & F1\_max & AUPRO & Params(M)      & FLOPs(G)      \\ \midrule
R224$^2$       & 99.3      & 99.8    & 98.8       & 98.2   & 60.8  & 61.9    & 93.6  & 139.8          & \textbf{32.3}          \\
R256$^2$-C224$^2$  & 99.3      & 99.8    & 99.0       & 98.1   & 64.2  & 64.4    & 92.7  & 139.8          & \textbf{32.3}          \\
R280$^2$       & 99.5      & \textbf{99.9}    & \textbf{99.2}       & 98.4   & 64.9  & 64.8    & 94.6  & 139.8          & 50.2          \\
R320$^2$-C280$^2$  & 99.6      & \textbf{99.9}    & 99.1       & 98.3   & 67.5  & 67.1    & 93.9  & 139.8          & 50.2          \\
R392$^2$       & 99.6      & 99.8    & 99.1       & \textbf{98.6}   & 69.1  & 68.5    & \textbf{95.6}  & 139.8          & 98.0          \\
\rowcolor{Light} 
R448$^2$-C392$^2$  & \textbf{99.7}      & \textbf{99.9}    & \textbf{99.2}       & 98.5   & \textbf{71.0}  & \textbf{69.7}    & 94.9  & 139.8          & 98.0          \\ \bottomrule[1.5pt]
\end{tabular}}}
\end{table*}
\begin{table}[]
\centering
\caption{Influence of the \textbf{Image Size} on the performance of other methods on MVTec-AD~\cite{MVTec-AD} dataset.}
\label{table:image_size_other}
\fontsize{11.5}{14}\selectfont{
\resizebox{\linewidth}{!}{
\begin{tabular}{c|c|cc}
\toprule[1.5pt]
Method                     & Input Size & Image-level    & Pixel-level         \\ \midrule
\multirow{3}{*}{RD4AD~\cite{RD4AD}}     & R256$^2$        & 94.6/96.5/96.1 & 96.1/48.6/53.8/91.1 \\
                           & R384$^2$         & 91.9/96.2/95.0 & 94.0/47.8/50.9/88.6 \\ \cmidrule{2-4} 
                           & $\triangle$          & \textcolor{purple}{-2.7/0.3/1.1}   & \textcolor{purple}{-2.1/0.8/2.9/2.5}    \\ \midrule
\multirow{3}{*}{SimpleNet~\cite{liu2023simplenet}} & R256$^2$         & 95.3/98.4/95.8 & 96.9/45.9/49.7/86.5 \\
                           & R384$^2$         & 86.1/93.6/90.9 & 89.5/36.0/40.5/76.4 \\ \cmidrule{2-4} 
                           & $\triangle$          & \textcolor{purple}{-9.2/4.8/4.9}   & \textcolor{purple}{-7.4/9.9/9.2/10.1}   \\ \midrule
\multirow{3}{*}{PatchCore~\cite{Patchcore}} & R256$^2$         & 97.2/99.1/97.2 & 97.9/53.8/56.3/91.3 \\
                           & R384$^2$         & 98.9/99.6/98.3 & 98.0/58.4/59.8/93.2 \\ \cmidrule{2-4} 
                           & $\triangle$          & \textcolor{purple}{+1.7/0.5/1.1}   & \textcolor{purple}{+0.1/4.6/3.5/1.9}    \\ \bottomrule[1.5pt]
\end{tabular}}}
\end{table}
\begin{table*}[!h]
\centering
\caption{Influence of the \textbf{ViT Architecture} on model performance for the MVTec-AD~\cite{MVTec-AD} dataset.}
\label{table:vit_architecture}
\fontsize{11.5}{14}\selectfont{
\resizebox{0.85\linewidth}{!}{
\begin{tabular}{c|ccc|cccc|cc}
\toprule[1.5pt]
Metric~$\rightarrow$       & \multicolumn{3}{c|}{Image-level} & \multicolumn{4}{c|}{Pixel-level} & \multicolumn{2}{c}{Efficiency} \\ \cmidrule{2-10} 
Architecture~$\downarrow$ & AUROC     & AP      & F1\_max    & AUROC  & AP    & F1\_max & AUPRO & Params(M)      & FLOPs(G)      \\ \midrule
ViT-Small    & 99.2      & 99.7    & 98.6       & 98.2   & 69.1  & 68.5    & 94.3  & \textbf{35.1}           & \textbf{24.6}          \\
\rowcolor{Light} 
ViT-Base     & 99.7      & \textbf{99.9}    & 99.2       & 98.5   & 71.0  & 69.7    & 94.9  & 139.8          & 98.0          \\
ViT-Large    & \textbf{99.8}      & \textbf{99.9}    & \textbf{99.4}       & \textbf{98.6}   & \textbf{72.1}  & \textbf{70.5}    & \textbf{95.6}  & 361.7          & 263.4         \\ \bottomrule[1.5pt]
\end{tabular}}}
\end{table*}

\begin{figure}
    \centering
    \includegraphics[width=0.9\linewidth]{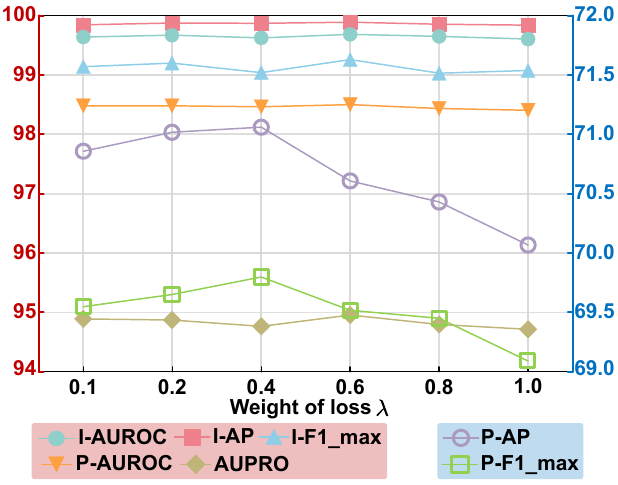}
    \caption{\textbf{Influence of the weight of loss function} $\lambda$ \textbf{on model performance for the MVTec-AD}~\cite{MVTec-AD} \textbf{dataset}. Pixel-
level AP and F1\_max use the \textcolor{blue}{right vertical axis}, while the other
metrics share the \textcolor{red}{left vertical axis}.}
    \label{fig:loss-weight}
\end{figure}

\begin{table*}[]
\centering
\caption{\textbf{Super-multi-class} anomaly detection performance on different AD datasets. $\Delta$ represents the performance change of INP-Former in the super-multi-class setting relative to the multi-class setting.}
\label{table:super-multi-class}
\fontsize{11.5}{14}\selectfont{
\resizebox{\linewidth}{!}{
\begin{tabular}{c|cccccc}
\toprule[1.5pt]
Dataset~$\rightarrow$           & \multicolumn{2}{c|}{MVTec-AD~\cite{MVTec-AD}}                             & \multicolumn{2}{c|}{VisA~\cite{VisA}}                                 & \multicolumn{2}{c}{Real-IAD~\cite{real-iad}}         \\ \midrule
Metric~$\rightarrow$            & \multicolumn{6}{c}{Image-level(I-AUROC/I-AP/I-F1\_max)\hspace{10mm} Pixel-level(P-AUROC/P-AP/P-F1\_max/AUPRO)}                                                                                                                  \\ \cmidrule{2-7} 
Setting~$\downarrow$           & Image-level    & \multicolumn{1}{c|}{Pixel-level}         & Image-level    & \multicolumn{1}{c|}{Pixel-level}         & Image-level    & Pixel-level         \\ \midrule
Multi-Class       & 99.7/99.9/99.2 & \multicolumn{1}{c|}{98.5/71.0/69.7/94.9} & 98.9/99.0/96.6 & \multicolumn{1}{c|}{98.9/51.2/54.7/94.4} & 90.5/88.1/81.5 & 99.0/47.5/50.3/95.0 \\
Super-Multi-Class & 99.5/99.8/98.9 & \multicolumn{1}{c|}{98.1/69.2/68.1/94.2} & 97.3/97.8/94.1 & \multicolumn{1}{c|}{98.4/51.4/54.7/92.4} & 89.8/87.4/80.5 & 98.9/45.2/48.6/94.4 \\ \midrule
$\Delta$                 & \textcolor{purple}{0.2$\downarrow$/0.1$\downarrow$/0.3$\downarrow$}    & \multicolumn{1}{c|}{\textcolor{purple}{0.4$\downarrow$/1.8$\downarrow$/1.6$\downarrow$/0.7$\downarrow$}}     & \textcolor{purple}{1.6$\downarrow$/1.2$\downarrow$/2.5$\downarrow$}    & \multicolumn{1}{c|}{\textcolor{purple}{0.5$\downarrow$/0.2$\uparrow$/0.0/2.0$\downarrow$}}   & \textcolor{purple}{0.7$\downarrow$/0.7$\downarrow$/1.0$\downarrow$}    & \textcolor{purple}{0.1$\downarrow$/2.3$\downarrow$/1.7$\downarrow$/0.6$\downarrow$}     \\ \bottomrule[1.5pt]
\end{tabular}}}
\end{table*}

\begin{table*}[]
\centering
\caption{\textbf{Few-shot (1-shot)} anomaly detection performance on different AD datasets. The best in \textbf{bold}, the
second-highest is \underline{underlined}. $\dagger$ indicates the results we reproduced using publicly available code.}
\label{table:1-shot-performance}
\fontsize{11}{14}\selectfont{
\resizebox{\linewidth}{!}{
\begin{tabular}{c|cccccc}
\toprule[1.5pt]
Dataset~$\rightarrow$             & \multicolumn{2}{c|}{MVTec-AD~\cite{MVTec-AD}}                                                     & \multicolumn{2}{c|}{VisA~\cite{VisA}}                                                         & \multicolumn{2}{c}{Real-IAD~\cite{real-iad}}         \\ \midrule
Metric~$\rightarrow$              & \multicolumn{6}{c}{Image-level(I-AUROC/I-AP/I-F1\_max)\hspace{10mm}Pixel-level(P-AUROC/P-AP/P-F1\_max/AUPRO)}                                                                                                                                                                  \\ \cmidrule{2-7} 
Method~$\downarrow$              & Image-level    & \multicolumn{1}{c|}{Pixel-level}                                 & Image-level    & \multicolumn{1}{c|}{Pixel-level}                                 & Image-level    & Pixel-level         \\ \midrule
SPADE~\cite{SPADE}               & 82.9/91.7/91.1 & \multicolumn{1}{c|}{92.0/-/44.5/85.7}                            & 79.5/82.0/80.7 & \multicolumn{1}{c|}{95.6/-/35.5/84.1}                            & 51.2$^\dagger$/45.6$^\dagger$/61.4$^\dagger$ & 59.5$^\dagger$/0.2$^\dagger$/0.5$^\dagger$/19.3$^\dagger$   \\
PaDiM~\cite{defard2021padim}               & 78.9/89.3/89.2 & \multicolumn{1}{c|}{91.3/-/43.7/78.2}                            & 62.8/68.3/75.3 & \multicolumn{1}{c|}{89.9/-/17.4/64.3}                            & 52.9$^\dagger$/47.4$^\dagger$/62.0$^\dagger$    & 84.9$^\dagger$/0.8$^\dagger$/2.3$^\dagger$/52.7$^\dagger$         \\
PatchCore~\cite{Patchcore}           & 86.3/93.8/92.0 & \multicolumn{1}{c|}{93.3/-/53.0/82.3}                            & 79.9/82.8/81.7 & \multicolumn{1}{c|}{95.4/-/38.0/80.5}                            & 59.3$^\dagger$/55.8$^\dagger$/\underline{62.3}$^\dagger$ & 89.6$^\dagger$/6.6$^\dagger$/12.3$^\dagger$/60.5$^\dagger$  \\
WinCLIP~\cite{WinClip}             & 93.1/\underline{96.5}/\underline{93.7} & \multicolumn{1}{c|}{95.2/-/\underline{55.9}/87.1}                            & 83.8/\underline{85.1}/\underline{83.1} & \multicolumn{1}{c|}{\underline{96.4}/-/\underline{41.3}/85.1}                            & \textbf{69.4}$^\dagger$/\underline{56.8}$^\dagger$/58.8$^\dagger$ & \underline{91.9}$^\dagger$/\underline{9.0}$^\dagger$/\underline{15.3}$^\dagger$/\underline{71.0}$^\dagger$  \\
PromptAD~\cite{PromptAD}            & \underline{94.6}/-/-       & \multicolumn{1}{c|}{\underline{95.9}/-/-/\underline{87.9}}                               & \underline{86.9}/-/-       & \multicolumn{1}{c|}{\textbf{96.7}/-/-/\underline{85.8}}                               & 52.2$^\dagger$/41.6$^\dagger$/52.2$^\dagger$ & 84.9$^\dagger$/7.6$^\dagger$/14.6$^\dagger$/58.4$^\dagger$  \\ \midrule
\rowcolor{Light} 
\textbf{INP-Former} & \textbf{96.6}/\textbf{98.2}/\textbf{96.4} & \multicolumn{1}{c|}{\cellcolor{Light}\textbf{97.0}/\textbf{64.2}/\textbf{64.0}/\textbf{92.6}} & \textbf{91.4}/\textbf{92.2}/\textbf{88.6} & \multicolumn{1}{c|}{\cellcolor{Light}96.3/\textbf{42.5}/\textbf{47.3}/\textbf{89.5}} & \underline{67.5}/\textbf{63.1}/\textbf{66.1} & \textbf{94.9}/\textbf{20.0}/\textbf{25.8}/\textbf{81.8} \\ \bottomrule[1.5pt]
\end{tabular}}}
\end{table*}

\begin{table*}[]
\centering
\caption{\textbf{Few-shot (2-shot)} anomaly detection performance on different AD datasets. The best in \textbf{bold}, the
second-highest is \underline{underlined}. $\dagger$ indicates the results we reproduced using publicly available code.}
\label{table:2-shot-performance}
\fontsize{11}{14}\selectfont{
\resizebox{\linewidth}{!}{
\begin{tabular}{c|cccccc}
\toprule[1.5pt]
Dataset~$\rightarrow$             & \multicolumn{2}{c|}{MVTec-AD~\cite{MVTec-AD}}                                                     & \multicolumn{2}{c|}{VisA~\cite{VisA}}                                                         & \multicolumn{2}{c}{Real-IAD~\cite{real-iad}}         \\ \midrule
Metric~$\rightarrow$              & \multicolumn{6}{c}{Image-level(I-AUROC/I-AP/I-F1\_max)\hspace{10mm}Pixel-level(P-AUROC/P-AP/P-F1\_max/AUPRO)}                                                                                                                                                                  \\ \cmidrule{2-7} 
Method~$\downarrow$              & Image-level    & \multicolumn{1}{c|}{Pixel-level}                                 & Image-level    & \multicolumn{1}{c|}{Pixel-level}                                 & Image-level    & Pixel-level         \\ \midrule
SPADE~\cite{SPADE}               & 81.0/90.6/90.3 & \multicolumn{1}{c|}{91.2/-/42.4/83.9}                            & 81.7/83.4/82.1 & \multicolumn{1}{c|}{96.2/-/40.5/85.7}                            & 50.9$^{\dagger}$/45.5$^{\dagger}$/61.2$^{\dagger}$ & 59.5$^{\dagger}$/0.2$^{\dagger}$/0.5$^{\dagger}$/19.2$^{\dagger}$   \\
PaDiM~\cite{defard2021padim}               & 76.6/88.1/88.2 & \multicolumn{1}{c|}{89.3/-/40.2/73.3}                            & 67.4/71.6/75.7 & \multicolumn{1}{c|}{92.0/-/21.1/70.1}                            & 55.9$^{\dagger}$/49.6$^{\dagger}$/62.9$^{\dagger}$ & 88.5$^{\dagger}$/1.5$^{\dagger}$/3.8$^{\dagger}$/61.6$^{\dagger}$   \\
PatchCore~\cite{Patchcore}           & 83.4/92.2/90.5 & \multicolumn{1}{c|}{92.0/-/\underline{58.4}/79.7}                            & 81.6/84.8/82.5 & \multicolumn{1}{c|}{96.1/-/41.0/82.6}                            & 63.3$^{\dagger}$/\underline{59.7}$^{\dagger}$/\underline{64.2}$^{\dagger}$ & 92.0$^{\dagger}$/9.4$^{\dagger}$/14.1$^{\dagger}$/66.1$^{\dagger}$  \\
WinCLIP~\cite{WinClip}             & 94.4/\underline{97.0}/\underline{94.4} & \multicolumn{1}{c|}{96.0/-/\underline{58.4}/88.4}                            & 84.6/\underline{85.8}/\underline{83.0} & \multicolumn{1}{c|}{96.8/-/\underline{43.5}/\underline{86.2}}                            & \textbf{70.9}$^{\dagger}$/58.7$^{\dagger}$/60.3$^{\dagger}$ & \underline{93.2}$^{\dagger}$/\underline{11.7}$^{\dagger}$/\underline{18.3}$^{\dagger}$/\underline{74.7}$^{\dagger}$ \\
PromptAD~\cite{PromptAD}            & \underline{95.7}/-/-       & \multicolumn{1}{c|}{\underline{96.2}/-/-/\underline{88.5}}                               & \underline{88.3}/-/-       & \multicolumn{1}{c|}{\underline{97.1}/-/-/85.8}                               & 57.7$^{\dagger}$/41.1$^{\dagger}$/52.9$^{\dagger}$ & 86.4$^{\dagger}$/8.5$^{\dagger}$/16.2$^{\dagger}$/61.0$^{\dagger}$  \\ \midrule
\rowcolor{Light} 
\textbf{INP-Former} & \textbf{97.0}/\textbf{98.2}/\textbf{96.7} & \multicolumn{1}{c|}{\cellcolor{Light}\textbf{97.2}/\textbf{66.0}/\textbf{65.6}/\textbf{93.1}} & \textbf{94.6}/\textbf{94.9}/\textbf{90.8} & \multicolumn{1}{c|}{\cellcolor{Light}\textbf{97.2}/\textbf{45.0}/\textbf{50.4}/\textbf{91.8}} & \underline{70.6}/\textbf{66.1}/\textbf{69.3} & \textbf{96.0}/\textbf{23.8}/\textbf{28.3}/\textbf{83.8} \\ \bottomrule[1.5pt]
\end{tabular}}}
\end{table*}

\begin{table*}[]
\centering
\caption{\textbf{Zero-shot} anomaly detection performance on different AD datasets. The best in \textbf{bold}.}
\label{table:zero-shot-performance}
\fontsize{10}{14}\selectfont{
\resizebox{0.85\linewidth}{!}{
\begin{tabular}{c|cccc}
\toprule[1.5pt]
Dataset~$\rightarrow$    & \multicolumn{2}{c|}{MVTec-AD~\cite{MVTec-AD}}                                                     & \multicolumn{2}{c}{VisA~\cite{VisA}}            \\ \midrule
Metric~$\rightarrow$     & \multicolumn{4}{c}{Image-level(I-AUROC/I-AP/I-F1\_max) Pixel-level(P-AUROC/P-AP/P-F1\_max/AUPRO)}                                                                             \\ \cmidrule{2-5} 
Method~$\downarrow$     & Image-level    & \multicolumn{1}{c|}{Pixel-level}                                 & Image-level    & Pixel-level        \\ \midrule
WinCLIP~\cite{WinClip}    & \textbf{91.8}/\textbf{96.5}/\textbf{92.9} & \multicolumn{1}{c|}{85.1/-/31.7/64.6}                            & \textbf{78.1}/\textbf{81.2}/\textbf{79.0} & 79.6/-/\textbf{14.8}/56.8   \\ 
\rowcolor{Light} 
\textbf{INP-Former} & 80.8/90.7/89.1 & \multicolumn{1}{c|}{\cellcolor{Light}\textbf{88.0}/\textbf{36.1}/\textbf{39.5}/\textbf{76.9}} & 67.5/71.6/75.0 & \textbf{88.7}/\textbf{7.8}/11.8/\textbf{67.2} \\ \bottomrule[1.5pt]
\end{tabular}}}
\end{table*}

\section{More details about comparison methods}\label{sec:appen_more_comparison_details}
The detailed information of the other compared methods in the experiment are as follows. Unless otherwise indicated, we utilize the performance metrics as reported in the original paper. In the few-shot setting on the Real-IAD~\cite{real-iad} dataset, SPADE~\cite{SPADE}~\footnote{\href{https://github.com/byungjae89/SPADE-pytorch}{https://github.com/byungjae89/SPADE-pytorch}},
PaDiM~\cite{defard2021padim}~\footnote{\href{https://github.com/xiahaifeng1995/PaDiM-Anomaly-Detection-Localization-master}{https://github.com/xiahaifeng1995/PaDiM-Anomaly-Detection-Localization-master}}, PatchCore~\cite{Patchcore}~\footnote{\href{https://github.com/hcw-00/PatchCore_anomaly_detection}{{https://github.com/hcw-00/PatchCore\_anomaly\_detection}}}, WinCLIP~\cite{WinClip}~\footnote{\href{https://github.com/zqhang/Accurate-WinCLIP-pytorch}{https://github.com/zqhang/Accurate-WinCLIP-pytorch}}, and PromptAD~\cite{PromptAD}~\footnote{\href{https://github.com/FuNz-0/PromptAD}{https://github.com/FuNz-0/PromptAD}} are run with the publicly available implementations.

RD4AD~\cite{RD4AD}: RD4AD is a robust baseline model for anomaly detection methods based on knowledge distillation and has been widely adopted by subsequent researchers.

UniAD~\cite{uniad}: UniAD is a baseline model for multi-class anomaly detection, which employs a Transformer-based non-identical mapping reconstruction model to enable complex multi-class semantic learning. Similarly,

SimpleNet~\cite{liu2023simplenet}: SimpleNet is an efficient and user-friendly network for anomaly detection and localization, which relies on a binary discriminator of adapted features to distinguish between anomalies and normal samples. 

DeSTSeg~\cite{zhang2023destseg}: DeSTSeg is an improved student-teacher framework for visual anomaly detection, integrating a denoising encoder-decoder and a segmentation network. 

DiAD~\cite{diad}: DiAD is a diffusion-based framework for multi-class anomaly detection, which incorporates the Semantic Guided network to recover anomalies while preserving semantics.

MambaAD~\cite{he2024mambaad}: MambaAD is a recently developed multi-class anomaly detection model with a Mamba decoder and locality-enhanced state space module, which captures long-range and local information effectively.

Dinomaly~\cite{guo2024dinomaly}: Dinomaly is a streamlined reverse distillation framework that employs linear attention mechanisms and loose reconstruction to achieve substantial performance gains.

SPADE~\cite{SPADE}: SPADE is an early anomaly detection method that aligns anomalous images with normal images using a multi-resolution feature pyramid.

PaDiM~\cite{defard2021padim}: PaDiM utilizes the pre-trained CNN features of normal samples to fit multivariate Gaussian distributions, which is a widely used baseline model.

Patchcore~\cite{Patchcore}: PatchCore is an important milestone approach. It utilizes a memory bank of core set sampled nominal patch features.

WinCLIP~\cite{WinClip}: WinCLIP introduces the first VLM-driven approach for zero-shot anomaly detection. It meticulously crafts a comprehensive suite of custom text prompts, optimized for identifying anomalies, and integrates a window scaling technique to achieve anomaly segmentation.


PromptAD~\cite{PromptAD}: PromptAD improves few-shot anomaly detection by automating prompt learning for one-class settings. It employs semantic concatenation to generate anomaly prompts and introduces an explicit margin.


\section{Complexity Comparisons}\label{sec:appen_complexity_comparisons}
Tab.~\ref{table:complexity} compares the proposed INP-Former with seven SOTA methods in terms of model size and computational complexity. 
Notably, our method's FLOPs are lower than those of DeSTSeg, DiAD, and Dinomaly, while its performance significantly exceeds theirs. Although our method has a larger parameter size and FLOPs than SimpleNet, UniAD, and MambaAD, it demonstrates a substantial improvement in detection performance. Furthermore, our approach is applicable to multi-class, few-shot, and single-class anomaly detection settings. It is noteworthy that we also report the efficiency and performance of two additional variants of INP-Former (INP-Former-S and INP-Former-S$^*$). INP-Former-S achieves a significant reduction in both parameters and FLOPs, with only a minor performance decline of \textcolor{purple}{0.4$\downarrow$}. \textbf{Even more remarkably, INP-Former-S* not only reduces FLOPs compared to MambaAD but also outperforms MambaAD \textcolor{purple}{1.1$\uparrow$} in terms of performance.} Overall, our method shows significant potential in industrial applications.

\section{Influence of Input Resolution}\label{sec:appen_influence_of_resolution}
As shown in Tab.~\ref{table:image_size}, we conducted an ablation study to evaluate the impact of input resolution on model performance. The results demonstrate that our method is robust to variations in image size for image-level anomaly detection. However, the image size has a slight effect on pixel-level anomaly localization performance. This is attributed to the patch size of 14 in the ViT, which results in smaller feature maps when the input image is reduced in size, leading to performance degradation. Therefore, in our study, we default to resizing the image to 448$\times$448 and then applying a center crop to 392$\times$392. Additionally, it is noteworthy that, under the R256$^2$-C224$^2$ setting, our method still achieves superior detection and localization performance compared to previous SOTA methods. Additionally, we analyze the effect of input size on the performance of other methods. As shown in Tab.~\ref{table:image_size_other}, we observe that not all models show improved performance with larger input sizes. For instance, when the input size is increased from 256 to 384, the performance of RD4AD and SimpleNet drops significantly. In contrast, our method consistently demonstrates superior detection performance across various input sizes, further validating the effectiveness of our approach.

\section{Influence of ViT Architectures.}\label{sec:appen_influence_of_backbone}
Tab.~\ref{table:vit_architecture} illustrates the effect of the ViT architecture on model performance. Our method demonstrates strong detection performance even with ViT-Small, with performance further improving as the ViT model size increases. Although ViT-Large achieves the best performance, its high FLOPs and parameter count make it less practical. Therefore, we default to using ViT-Base in this study.

\section{Influence of the Weight of Loss Functions}\label{sec:appen_influence_of_loss_weight}
Fig.~\ref{fig:loss-weight} illustrates the effect of the weight of loss function on model performance in the MVTec-AD~\cite{MVTec-AD} dataset. Our method shows strong robustness to changes in weight of loss function at the image level. However, pixel-level performance initially increases and then decreases as the $\lambda$ grows. This trend occurs because, when $\lambda$ is too low, the INP Extractor may fail to consistently capture normal patterns, potentially including some anomalous information. Conversely, when $\lambda$ is too high, the model focuses excessively on updating the INP Extractor, overlooking updates to the INP-Guided Decoder, which leads to insufficient detail in reconstructed features. Based on these observations, we set $\lambda$ to 0.2 in our study.

\begin{figure}[!t]
    \centering
    \includegraphics[width=\linewidth]{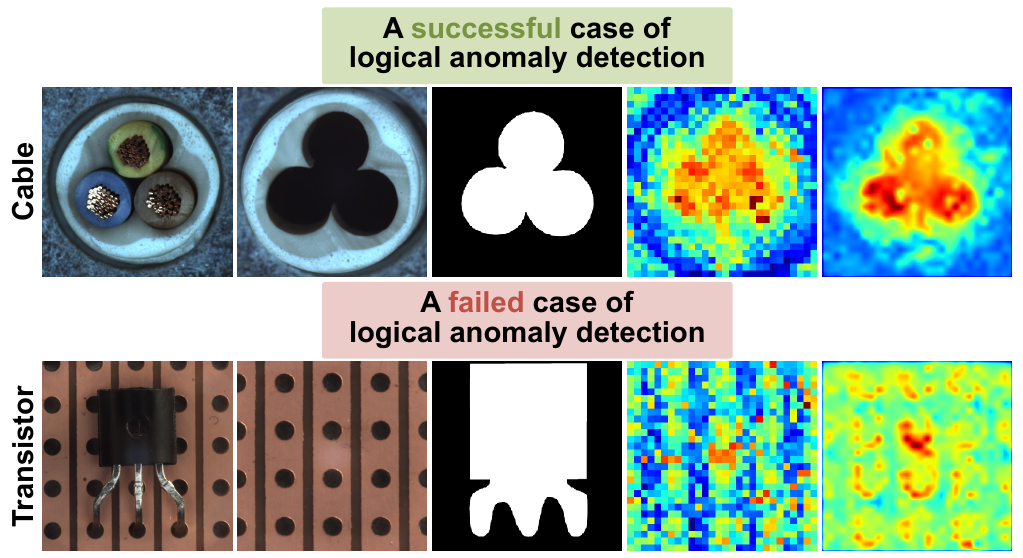}
    \caption{\textbf{Limitation of proposed method in detecting logical anomalies similar to the background}. 
From left to right: Normal Image, Input Anomaly, Ground Truth, Distance Map, and Predicted Anomaly Map.}
    \label{fig:limitation}
\end{figure}
\section{Super-Multi-Class Anomaly Detection}\label{sec:appen_super_multi_class_anomaly_detection_results}
Tab.~\ref{table:super-multi-class} presents the super-multi-class anomaly detection performance of INP-Former, \ie, training together with MVTec-AD, VisA, and Real-IAD. Compared to the multi-class anomaly detection setting, the performance of INP-Former in the super-multi-class setting only slightly declines. This demonstrates that our method can utilize a unified model to detect a broader range of products, which can significantly reduce memory consumption in industrial applications.
\section{Per-Class Multi-Class Anomaly Detection Results}\label{sec:appen_per_class_multi_class_anomaly_detection_results}
In this section, we present the performance of each class on the MVTec-AD~\cite{MVTec-AD}, VisA~\cite{VisA}, and Real-IAD~\cite{real-iad} datasets for multi-class anomaly detection. The performance of the comparison methods is derived from MambaAD~\cite{he2024mambaad} and Dinomaly~\cite{guo2024dinomaly}. Tab.~\ref{tab:mvtec_image_per_class} and Tab.~\ref{tab:mvtec_pixel_per_class} provide the results for image-level anomaly detection and pixel-level anomaly localization on the MVTec-AD dataset, respectively. Tab.~\ref{tab:visa_image_per_class} and Tab.~\ref{tab:visa_pixel_per_class} further present the corresponding results on the VisA dataset. Tab.~\ref{tab:Real_IAD_Image_per_class} and Tab.~\ref{tab:Real_IAD_Pixel_per_class} display the results for image-level anomaly detection and pixel-level anomaly localization on the Real-IAD dataset. These results convincingly demonstrate the superiority of our proposed method.

\section{More Few-shot Anomaly Detection Results}\label{sec:appen_per_class_few_shot_anomaly_detection_results}
Tab.~\ref{table:1-shot-performance} and Tab.~\ref{table:2-shot-performance} show the performance comparison between our method and existing methods across three datasets under 1-shot and 2-shot anomaly detection settings, respectively. Our method achieves state-of-the-art or competitive results across all three datasets, highlighting its superior effectiveness.

\section{Per-Class Single-Class Anomaly Detection Results}\label{sec:appen_per_class_single_class_anomaly_detection_results}
To support future research, we report the per-class performance of INP-Former in the single-class anomaly detection setting on MVTec-AD~\cite{MVTec-AD}, VisA~\cite{VisA}, and Real-IAD~\cite{real-iad} datasets. in Tab.~\ref{table:singleclass-mvtec}, Tab.~\ref{table:singleclass-visa}, and Tab.~\ref{table:singleclass-real-iad}, respectively.

\section{More Zero-shot Anomaly Detection Results}\label{sec:appen_per_class_zero_shot_anomaly_detection_results}
Tab.~\ref{table:zero-shot-performance} compares the zero-shot anomaly detection performance of our method with WinCLIP~\cite{WinClip}, a method specifically designed for zero-shot anomaly detection. Notably, we utilize INP-Former to extract INPs for images from unseen classes and then directly compare all tokens to these INPs for zero-shot anomaly detection. Although our method is not designed for zero-shot anomaly detection, it still possesses some efficacy for this task, with 88.0 and 88.7 pixel-level AUROCs on MVTec-AD and VisA, respectively. In terms of image-level performance, our method performs weaker than the existing specified method. We believe incorporating INPs with other specified designs can bring better zero-shot anomaly detection performance.

\section{More qualitative results}\label{sec:appen_more_qualitative_results}
\begin{figure*}
    \centering
    \includegraphics[width=\textwidth]{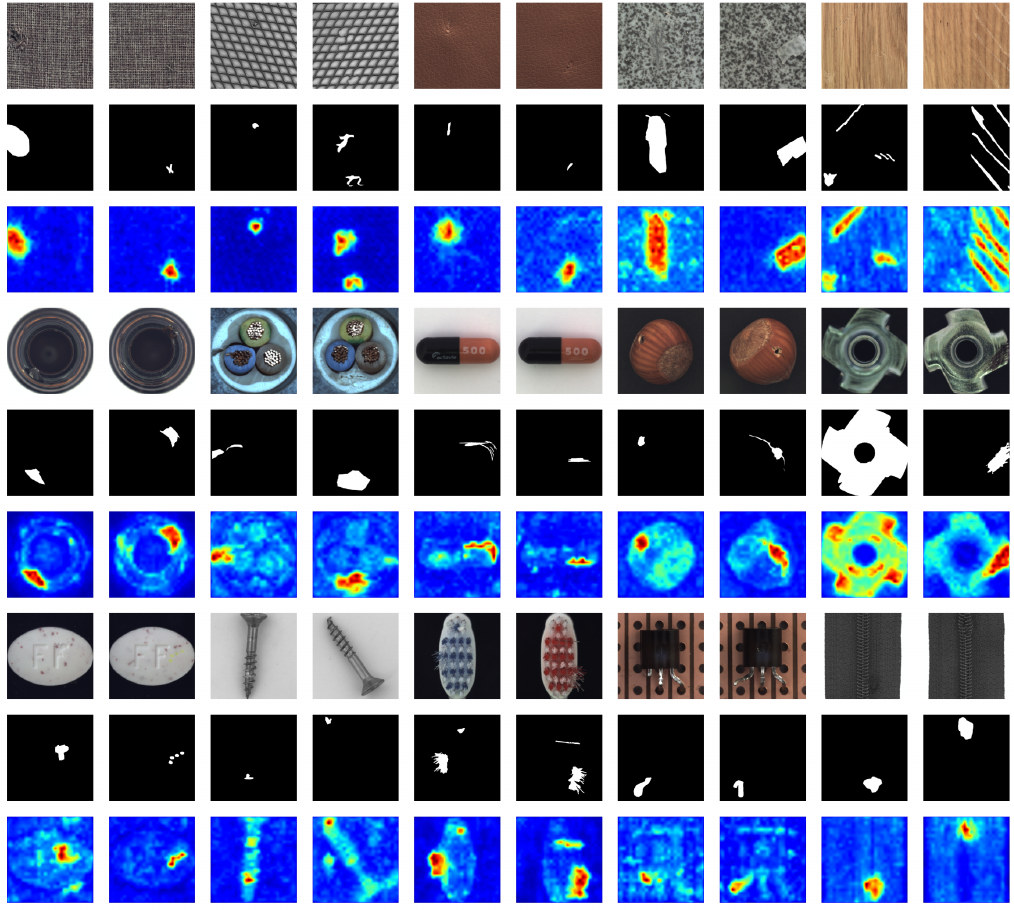}
    \caption{\textbf{Anomaly localization results on the MVTec-AD}~\cite{MVTec-AD} \textbf{dataset under the multi-class anomaly detection setting}. For each tuple, the images from top to bottom represent the anomaly image, ground truth, and predicted anomaly map.}
    \label{fig:mvtec-vis}
\end{figure*}

\begin{figure*}
    \centering
    \includegraphics[width=\textwidth]{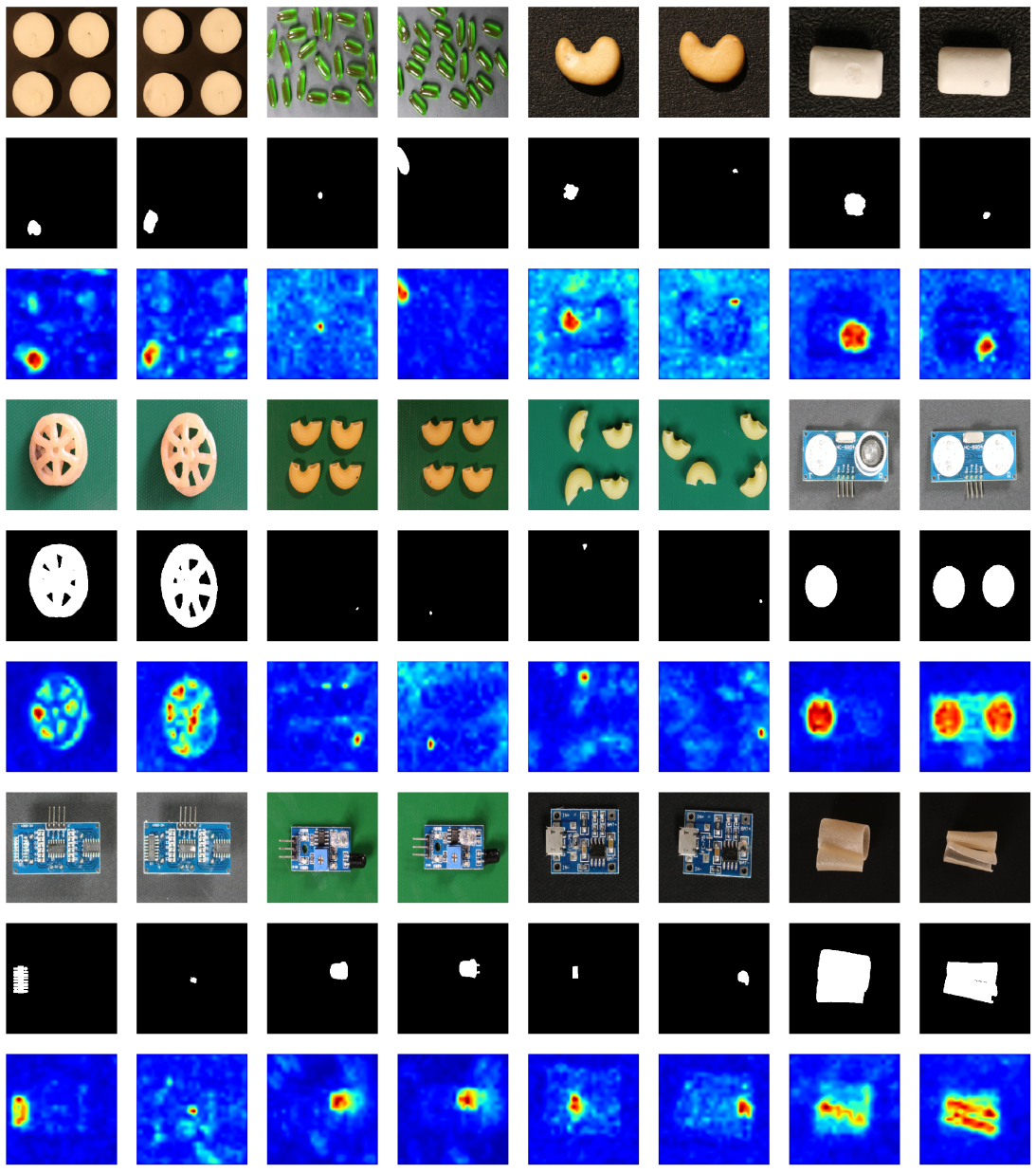}
    \caption{\textbf{Anomaly localization results on the VisA}~\cite{VisA} \textbf{dataset under the multi-class anomaly detection setting}. For each tuple, the images from top to bottom represent the anomaly image, ground truth, and predicted anomaly map.}
    \label{fig:visa-vis}
\end{figure*}

\begin{figure*}
    \centering
    \includegraphics[width=0.75\textwidth]{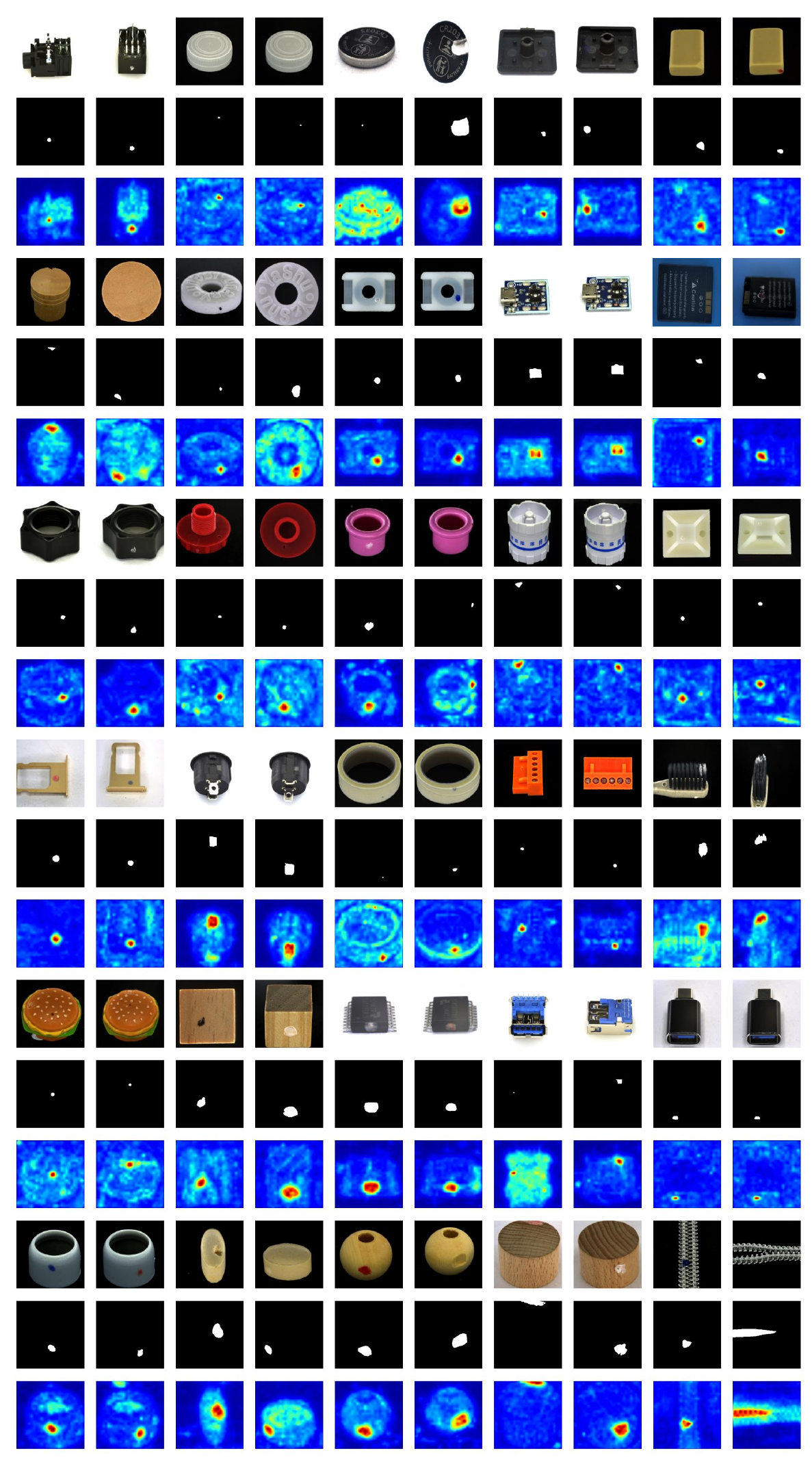}
    \caption{\textbf{Anomaly localization results on the Real-IAD}~\cite{real-iad} \textbf{dataset under the multi-class anomaly detection setting}. For each tuple, the images from top to bottom represent the anomaly image, ground truth, and predicted anomaly map.}
    \label{fig:real-iad-vis}
\end{figure*}

Fig.~\ref{fig:mvtec-vis}, Fig.~\ref{fig:visa-vis}, and Fig.~\ref{fig:real-iad-vis} display the predicted anomaly maps of our method on the MVTec-AD~\cite{MVTec-AD}, VisA~\cite{VisA}, and Real-IAD~\cite{real-iad} datasets for multi-class anomaly detection. These results clearly indicate that our approach can accurately localize anomalous regions for a wide range of categories.

\section{More detailed analysis of the limitations}
\label{sec:appen_more_limitations}
Fig.~\ref{fig:limitation} illustrates two examples of logical anomaly detection using our method. Interestingly, the misplaced logical anomaly in Cable is successfully detected, while the misplaced anomaly in Transistor is completely missed. We hypothesize that this is due to the significant difference between the misplaced anomaly and the background in Cable, whereas the misplaced anomaly in Transistor closely resembles the background. As a result, the INP Extractor mistakenly extracts the misplaced anomaly in Transistor as INPs, leading to a missed detection. This highlights a limitation of our method when dealing with logical anomalies that are similar to the background. In future work, we aim to combine pre-stored prototypes with INPs to address this issue. Pre-stored prototypes capture comprehensive semantic information, while INPs exhibit strong alignment. The integration of both is expected to improve the model’s performance in detecting logical anomalies that resemble the background.

\section{Comparison of INP with handcrafted aggregated prototypes}
\label{sec:compared_with_r1}
Although the concept in Reference~\cite{INPtexture} is similar to our proposed INP, we wish to emphasize that our method is fundamentally distinct. Reference~\cite{INPtexture} manually aggregates features within a single image as prototypes, and its scope is limited to zero-shot texture anomaly detection. In contrast, we introduce a learnable INP extractor that extracts normal features with adaptable shapes as INPs. This enables our method to be applied not only to textures but also to objects. Additionally, we integrate the INP into a reconstruction framework by proposing an INP-guided decoder, which not only reduces the computational cost of self-attention but also achieves superior detection performance across multiple settings.
\section{Comparision of INP with MuSc}
\label{sec:compared_with_Musc}
It may seem unusual that MuSc~\cite{li2024musc} performs better in zero-shot settings compared to our INP-Former in few-shot settings. However, this difference stems from the distinct setups of the two methods. MuSc is specifically designed for zero-shot detection and relies on a large number of test images for mutual scoring. In contrast, our INP-Former only requires a single image during the testing phase, making it adaptable to various settings. As such, comparing our method with MuSc is not a fair comparison.

\section{More visualizations of INPs}
\label{sec:assumption}
\begin{figure*}[!ht]
    \centering
    \includegraphics[width=\linewidth]{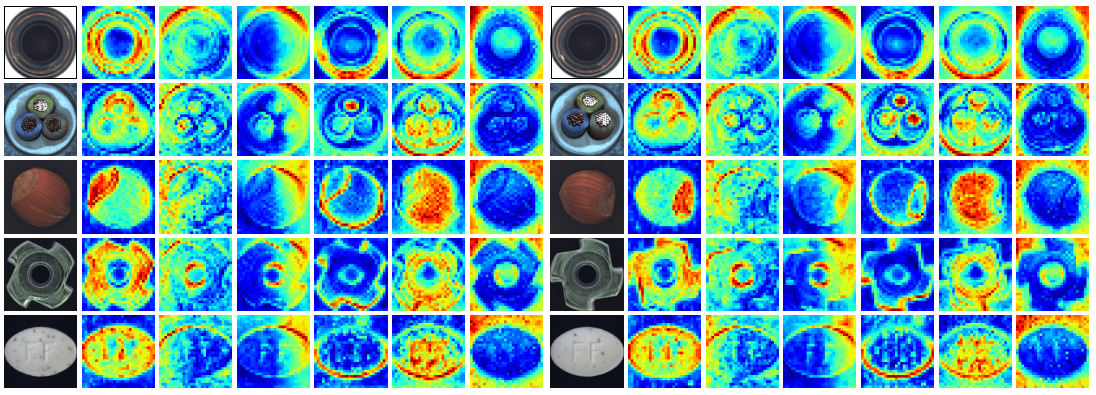}
    \caption{\textbf{Cross-attention maps} between INPs and image patches.}
    \label{fig:moreinpvis}
\end{figure*}

Fig.~\ref{fig:moreinpvis} presents the cross-attention maps between INPs and image patches. This clearly demonstrates that our INPs are able to capture semantic information from various regions, including object regions, object boundaries, and background areas.
\clearpage

\begin{table*}[]
\centering
\caption{\textbf{Per-Class Performance of the Proposed INP-Former on the MVTec-AD}~\cite{MVTec-AD} \textbf{Dataset for Single-Class Anomaly Detection}}
\label{table:singleclass-mvtec}
\fontsize{11}{14}\selectfont{
\resizebox{0.85\linewidth}{!}{
\begin{tabular}{c|ccc|cccc}
\toprule[1.5pt]
Metric~$\rightarrow$     & \multicolumn{3}{c|}{Image-level} & \multicolumn{4}{c}{Pixel-level}      \\ \cmidrule{2-8} 
Category~$\downarrow$   & I-AUROC   & I-AP   & I-F1\_max   & P-AUROC & P-AP & P-F1\_max & AUPRO \\ \midrule
Bottle     & 100       & 100    & 100         & 99.1    & 88.9 & 82.4      & 97.2    \\
Cable      & 100       & 100    & 100         & 98.8    & 78.9 & 75.0      & 95.2    \\
Capsule    & 98.6      & 99.7   & 98.2        & 98.5    & 60.1 & 57.5      & 97.7    \\
Hazelnut   & 100       & 100    & 100         & 99.5    & 82.8 & 78.4      & 97.0    \\
Metal Nut  & 100       & 100    & 100         & 97.1    & 81.1 & 86.3      & 94.3    \\
Pill       & 99.2      & 99.9   & 98.6        & 96.0    & 66.8 & 66.9      & 97.2    \\
Screw      & 98.0      & 99.3   & 96.7        & 99.6    & 63.8 & 59.9      & 98.3    \\
Toothbrush & 100       & 100    & 100         & 99.1    & 57.4 & 67.4      & 95.8    \\
Transistor & 99.9      & 99.8   & 98.8        & 95.6    & 66.7 & 63.7      & 86.4    \\
Zipper     & 100       & 100    & 100         & 98.1    & 71.6 & 68.3      & 94.6    \\
Carpet     & 99.9      & 99.9   & 99.4        & 99.4    & 75.7 & 72.6      & 98.0    \\
Grid       & 100       & 100    & 100         & 99.5    & 61.6 & 62.0      & 97.6    \\
Leather    & 100       & 100    & 100         & 99.3    & 52.2 & 53.3      & 98.4    \\
Tile       & 100       & 100    & 100         & 97.8    & 73.0 & 75.7      & 88.9    \\
Wood       & 99.8      & 99.9   & 99.2        & 97.4    & 72.1 & 68.4      & 94.1    \\ \midrule
Mean       & 99.7      & 99.9   & 99.4        & 98.3    & 70.2 & 69.2      & 95.4    \\ \bottomrule[1.5pt]
\end{tabular}}}
\end{table*}

\begin{table*}[h]
\centering
\caption{\textbf{Per-Class Performance of the Proposed INP-Former on the VisA}~\cite{VisA} \textbf{Dataset for Single-Class Anomaly Detection}}
\label{table:singleclass-visa}
\fontsize{11}{14}\selectfont{
\resizebox{0.85\linewidth}{!}{
\begin{tabular}{c|ccc|cccc}
\toprule[1.5pt]
Metric~$\rightarrow$      & \multicolumn{3}{c|}{Image-level} & \multicolumn{4}{c}{Pixel-level}      \\ \cmidrule{2-8} 
Category~$\downarrow$    & I-AUROC   & I-AP   & I-F1\_max   & P-AUROC & P-AP & P-F1\_max & AUPRO \\ \midrule
pcb1        & 98.6      & 98.5   & 95.0        & 99.6    & 86.7 & 78.5      & 95.1    \\
pcb2        & 98.0      & 96.6   & 96.0        & 98.8    & 40.0 & 40.3      & 91.1    \\
pcb3        & 99.3      & 99.4   & 97.0        & 99.0    & 28.4 & 38.5      & 93.1    \\
pcb4        & 100       & 100    & 99.0        & 98.6    & 51.3 & 51.6      & 93.2    \\
macaroni1   & 97.2      & 97.3   & 91.0        & 99.4    & 33.1 & 40.4      & 94.8    \\
macaroni2   & 95.0      & 94.9   & 89.0        & 99.7    & 26.7 & 36.2      & 98.4    \\
capsules    & 99.0      & 99.3   & 97.6        & 99.5    & 66.2 & 65.5      & 98.2    \\
candle      & 98.8      & 98.8   & 94.7        & 99.4    & 46.2 & 50.2      & 95.7    \\
cashew      & 98.5      & 99.3   & 97.0        & 93.8    & 59.9 & 60.6      & 89.7    \\
chewinggum  & 99.4      & 99.7   & 96.9        & 98.8    & 58.1 & 63.5      & 87.1    \\
fryum       & 99.3      & 99.7   & 98.0        & 95.8    & 43.5 & 48.8      & 93.1    \\
pipe\_fryum & 99.2      & 99.6   & 97.5        & 98.5    & 49.8 & 56.5      & 95.6    \\ \midrule
Mean        & 98.5      & 98.6   & 95.7        & 98.4    & 49.2 & 52.6      & 93.8    \\ \bottomrule[1.5pt]
\end{tabular}}}
\end{table*}

\begin{table*}[]
\centering
\caption{\textbf{Per-Class Performance of the Proposed INP-Former on the Real-IAD}~\cite{real-iad} \textbf{Dataset for Single-Class Anomaly Detection}}
\label{table:singleclass-real-iad}
\fontsize{11}{14}\selectfont{
\resizebox{0.85\linewidth}{!}{
\begin{tabular}{c|ccc|cccc}
\toprule[1.5pt]
Metric~$\rightarrow$              & \multicolumn{3}{c|}{Image-level} & \multicolumn{4}{c}{Pixel-level}      \\ \cmidrule{2-8} 
Category~$\downarrow$            & I-AUROC   & I-AP   & I-F1\_max   & P-AUROC & P-AP & P-F1\_max & AUPRO \\ \midrule
audiojack           & 92.2      & 88.2   & 78.4        & 99.6    & 53.3 & 55.2      & 97.0    \\
bottle\_cap         & 94.9      & 94.1   & 85.0        & 99.7    & 40.2 & 40.4      & 98.4    \\
button\_battery     & 89.9      & 91.4   & 84.4        & 99.2    & 51.9 & 56.3      & 93.8    \\
end\_cap            & 89.5      & 89.1   & 85.5        & 99.3    & 23.9 & 34.9      & 97.0    \\
eraser              & 93.9      & 91.9   & 83.1        & 99.8    & 48.2 & 50.7      & 98.3    \\
fire\_hood          & 88.4      & 81.7   & 73.3        & 99.6    & 47.2 & 49.6      & 96.5    \\
\end{tabular}}}
\end{table*}
\begin{table*}[!ht]
\centering
\caption{Per-Class Results on the \textbf{MVTec-AD}~\cite{MVTec-AD} Dataset for \textbf{Multi-Class Anomaly Detection} with AUROC/AP/F1\_max metrics.}
\label{tab:mvtec_image_per_class}
\fontsize{11.5}{14}\selectfont{
\resizebox{\textwidth}{!}{
\begin{tabular}{c|ccccccc|
>{\columncolor{Light}}c}
\toprule[1.5pt]
Method~$\rightarrow$   & RD4AD~\cite{RD4AD}                                                                            & UniAD~\cite{uniad}                                                                            & SimpleNet~\cite{liu2023simplenet}                                                                        & DeSTSeg~\cite{zhang2023destseg}                                                                               & DiAD~\cite{diad}                                                            & MambaAD~\cite{he2024mambaad}                                                                        & Dinomaly~\cite{guo2024dinomaly}                                                                                   & \textbf{INP-Former}                                                                                \\ 
Category~$\downarrow$ & CVPR'22                                                                         & NeurlPS'22                                                                       & CVPR'23                                                                          & CVPR'23                                                                               & AAAI'24                                                         & NeurIPS'24                                                                       & Arxiv'24                                                                                     & \textbf{Ours}                                                                                      \\
\end{tabular}}}
\end{table*}

\begin{table*}[!ht]
\centering
\caption{Per-Class Results on the \textbf{MVTec-AD}~\cite{MVTec-AD} Dataset for \textbf{Multi-Class Anomaly Localization} with AUROC/AP/F1\_max/AUPRO metrics.}
\label{tab:mvtec_pixel_per_class}
\fontsize{14}{18}\selectfont{
\resizebox{\textwidth}{!}{
\begin{tabular}{c|ccccccc|
>{\columncolor{Light}}c}
\toprule[1.5pt]
Method~$\rightarrow$   & RD4AD~\cite{RD4AD}                                                                            & UniAD~\cite{uniad}                                                                            & SimpleNet~\cite{liu2023simplenet}                                                                        & DeSTSeg~\cite{zhang2023destseg}                                                                               & DiAD~\cite{diad}                                                            & MambaAD~\cite{he2024mambaad}                                                                        & Dinomaly~\cite{guo2024dinomaly}                                                                                   & \textbf{INP-Former}                                                                                \\ 
Category~$\downarrow$ & CVPR'22                                                                         & NeurlPS'22                                                                       & CVPR'23                                                                          & CVPR'23                                                                               & AAAI'24                                                         & NeurIPS'24                                                                       & Arxiv'24                                                                                     & \textbf{Ours}                                                                                                                                           \\
\end{tabular}}}
\end{table*}

\begin{table*}[]
\centering
\caption{Per-Class Results on the \textbf{VisA}~\cite{VisA} Dataset for \textbf{Multi-Class Anomaly Detection} with AUROC/AP/F1\_max metrics.}
\label{tab:visa_image_per_class}
\fontsize{11.5}{14}\selectfont{
\resizebox{\textwidth}{!}{
\begin{tabular}{c|ccccccc|
>{\columncolor{Light}}c }
\toprule[1.5pt]
Method~$\rightarrow$   & RD4AD~\cite{RD4AD}                                                                            & UniAD~\cite{uniad}                                                                            & SimpleNet~\cite{liu2023simplenet}                                                                        & DeSTSeg~\cite{zhang2023destseg}                                                                               & DiAD~\cite{diad}                                                            & MambaAD~\cite{he2024mambaad}                                                                        & Dinomaly~\cite{guo2024dinomaly}                                                                                   & \textbf{INP-Former}                                                                                \\ 
Category~$\downarrow$ & CVPR'22                                                                         & NeurlPS'22                                                                       & CVPR'23                                                                          & CVPR'23                                                                               & AAAI'24                                                         & NeurIPS'24                                                                       & Arxiv'24                                                                                     & \textbf{Ours}                                                                                                          \\ \midrule
pcb1        & 96.2/95.5/91.9                                                                               & 92.8/92.7/87.8 & 91.6/91.9/86.0 & 87.6/83.1/83.7 & 88.1/88.7/80.7 & 95.4/93.0/91.6                                                     & \textbf{99.1}/\textbf{99.1}/\textbf{96.6}          & \underline{98.8}/\underline{98.7}/\underline{96.1} \\
\end{tabular}}}
\end{table*}

\begin{table*}[]
\centering
\caption{Per-Class Results on the \textbf{VisA}~\cite{VisA} Dataset for \textbf{Multi-Class Anomaly Localization} with AUROC/AP/F1\_max/AUPRO metrics.}
\label{tab:visa_pixel_per_class}
\fontsize{14}{18}\selectfont{
\resizebox{\textwidth}{!}{
\begin{tabular}{c|ccccccc|
>{\columncolor{Light}}c }
\toprule[1.5pt]
Method~$\rightarrow$   & RD4AD~\cite{RD4AD}                                                                            & UniAD~\cite{uniad}                                                                            & SimpleNet~\cite{liu2023simplenet}                                                                        & DeSTSeg~\cite{zhang2023destseg}                                                                               & DiAD~\cite{diad}                                                            & MambaAD~\cite{he2024mambaad}                                                                        & Dinomaly~\cite{guo2024dinomaly}                                                                                   & \textbf{INP-Former}                                                                                \\ 
Category~$\downarrow$ & CVPR'22                                                                         & NeurlPS'22                                                                       & CVPR'23                                                                          & CVPR'23                                                                               & AAAI'24                                                         & NeurIPS'24                                                                       & Arxiv'24                                                                                     & \textbf{Ours}                                                                                                                            \\
\end{tabular}}}
\end{table*}

\begin{table*}[]
\centering
\caption{Per-Class Results on the \textbf{Real-IAD}~\cite{real-iad} Dataset for \textbf{Multi-Class Anomaly Detection} with AUROC/AP/F1\_max metrics.}
\label{tab:Real_IAD_Image_per_class}
\fontsize{11.5}{14}\selectfont{
\resizebox{\textwidth}{!}{
\begin{tabular}{c|ccccccc|
>{\columncolor{Light}}c }
\toprule[1.5pt]
Method~$\rightarrow$   & RD4AD~\cite{RD4AD}                                                                            & UniAD~\cite{uniad}                                                                            & SimpleNet~\cite{liu2023simplenet}                                                                        & DeSTSeg~\cite{zhang2023destseg}                                                                               & DiAD~\cite{diad}                                                            & MambaAD~\cite{he2024mambaad}                                                                        & Dinomaly~\cite{guo2024dinomaly}                                                                                   & \textbf{INP-Former}                                                                                \\ 
Category~$\downarrow$ & CVPR'22                                                                         & NeurlPS'22                                                                       & CVPR'23                                                                          & CVPR'23                                                                               & AAAI'24                                                         & NeurIPS'24                                                                       & Arxiv'24                                                                                     & \textbf{Ours}                                                                                         \\
\end{tabular}}}
\end{table*}

\begin{table*}[]
\centering
\caption{Per-Class Results on the \textbf{Real-IAD}~\cite{real-iad} Dataset for \textbf{Multi-Class Anomaly Localization} with AUROC/AP/F1\_max/AUPRO metrics.}
\label{tab:Real_IAD_Pixel_per_class}
\fontsize{14}{18}\selectfont{
\resizebox{\textwidth}{!}{
\begin{tabular}{c|ccccccc|
>{\columncolor{Light}}c }
\toprule[1.5pt]
Method~$\rightarrow$   & RD4AD~\cite{RD4AD}                                                                            & UniAD~\cite{uniad}                                                                            & SimpleNet~\cite{liu2023simplenet}                                                                        & DeSTSeg~\cite{zhang2023destseg}                                                                               & DiAD~\cite{diad}                                                            & MambaAD~\cite{he2024mambaad}                                                                        & Dinomaly~\cite{guo2024dinomaly}                                                                                   & \textbf{INP-Former}                                                                                \\ 
Category~$\downarrow$ & CVPR'22                                                                         & NeurlPS'22                                                                       & CVPR'23                                                                          & CVPR'23                                                                               & AAAI'24                                                         & NeurIPS'24                                                                       & Arxiv'24                                                                                     & \textbf{Ours}                                                                                                                    \\
\end{tabular}}}
\end{table*}

\end{document}